\documentclass{article}

     \PassOptionsToPackage{numbers, compress}{natbib}

     \usepackage[final]{neurips_2021}

\usepackage[ruled,vlined]{algorithm2e}
\usepackage{algorithmic}
\usepackage[utf8]{inputenc} 
\usepackage[T1]{fontenc}    
\usepackage{hyperref}       
\usepackage{url}            
\usepackage{booktabs}       
\usepackage{amsfonts}       
\usepackage{nicefrac}       
\usepackage{microtype}      
\usepackage[table,dvipsnames]{xcolor}          
\usepackage{amsmath}
\usepackage{enumitem}
\usepackage{amsthm}
\usepackage{graphicx}
\usepackage{subcaption}
\usepackage{color, colortbl}
\usepackage{wrapfig}
\usepackage{makecell}
\usepackage{listings}
\usepackage{wrapfig}
\usepackage{nicefrac}
\usepackage{multirow}
\usepackage{pifont}
\usepackage{tabularx,ragged2e}

\newtheorem{theorem}{Theorem}
\newtheorem{lemma}[theorem]{Lemma}

\newcommand{\crypten}{\textsc{CrypTen}}

\newcommand{\bx}{\mathbf{x}}
\newcommand{\by}{\mathbf{y}}

\newcommand{\thefield}{\mathbb{Z}/Q\mathbb{Z}}

\DeclareFixedFont{\ttb}{T1}{txtt}{bx}{n}{7} 
\DeclareFixedFont{\ttm}{T1}{txtt}{m}{n}{7}  
\DeclareFixedFont{\ttmlarge}{T1}{txtt}{m}{n}{9}  
\definecolor{deepblue}{rgb}{0,0,0.5}
\definecolor{deepred}{rgb}{0.6,0,0}
\definecolor{deepgreen}{rgb}{0,0.5,0}

\newcommand\pythonstyle{\lstset{
language=Python,
basicstyle=\linespread{0.8}\selectfont\ttm,
commentstyle=\linespread{0.8}\selectfont\ttm\color{deepgreen},
otherkeywords={self},             
keywordstyle=\linespread{0.8}\selectfont\ttb\color{deepblue},
emph={incmatrix},          
emphstyle=\ttb\color{deepred},    
stringstyle=\color{deepgreen},
frame=tb,                         
showstringspaces=false            %
}}

\lstnewenvironment{python}[1][]
{
\pythonstyle
\lstset{#1}
}
{}

\newcommand\pythoninline[1]{{\pythonstyle\lstinline!#1!}}

\newcommand*\colorcheck[1]{%
  \expandafter\newcommand\csname #1check\endcsname{\textcolor{#1}{\ding{52}}}%
}
\colorcheck{green}
\colorcheck{ForestGreen}
\newcommand*\colorcross[1]{%
  \expandafter\newcommand\csname #1cross\endcsname{\textcolor{#1}{\ding{55}}}%
}
\colorcross{red}

\makeatletter
\@ifundefined{c@rownum}{%
  \let\c@rownum\rownum
}{}
\@ifundefined{therownum}{%
  \def\therownum{\@arabic\rownum}%
}{}
\makeatother

\title{\crypten{}: Secure Multi-Party Computation\\Meets Machine Learning}

\author{
  Brian Knott
  \quad\quad\quad\enskip Shobha Venkataraman
  \quad\quad\quad\enskip Awni Hannun\\
  \quad\quad\quad\enskip \bf Shubho Sengupta
  \quad\quad\quad\enskip \bf Mark Ibrahim
  \quad\quad\quad\enskip \bf Laurens van der Maaten\\
  Facebook AI Research\\
  \texttt{\{brianknott,shobha,awni,ssengupta,marksibrahim,lvdmaaten\}@fb.com} \\
}

\begin{document}

\maketitle

\begin{abstract}
Secure multi-party computation (MPC) allows parties to perform computations on data while keeping that data private.
This capability has great potential for machine-learning applications: it facilitates training of machine-learning models on private data sets owned by different parties, evaluation of one party's private model using another party's private data, \textit{etc.}
Although a range of studies implement machine-learning models via secure MPC, such implementations are not yet mainstream.
Adoption of secure MPC is hampered by the absence of flexible software frameworks that ``speak the language'' of machine-learning researchers and engineers.
To foster adoption of secure MPC in machine learning, we present \crypten{}: a software framework that exposes popular secure MPC primitives via abstractions that are common in modern machine-learning frameworks, such as tensor computations, automatic differentiation, and modular neural networks.
This paper describes the design of \crypten{} and measure its performance on state-of-the-art models for text classification, speech recognition, and image classification.
Our benchmarks show that \crypten{}'s GPU support and high-performance communication between (an arbitrary number of) parties allows it to perform efficient private evaluation of modern machine-learning models under a \emph{semi-honest} threat model.
For example, two parties using \crypten{} can securely predict phonemes in speech recordings using Wav2Letter~\cite{collobert2016} faster than real-time.
We hope that \crypten{} will spur adoption of secure MPC in the machine-learning community.
\end{abstract}
\section{Introduction}
\label{sec:introduction}
Secure multi-party computation (MPC;~\cite{evans2018,yao86}) allows parties to collaboratively perform computations on their combined data sets without revealing the data they possess to each other.
This capability of secure MPC has the potential to unlock a variety of machine-learning applications that are currently infeasible because of data privacy concerns.
For example, secure MPC can allow medical research institutions to jointly train better diagnostic models without having to share their sensitive patient data~\cite{dugan2016} or allow social scientists to analyze gender wage gap statistics without companies having to share sensitive salary data~\cite{lapets2016}.
The prospect of such applications of machine learning with rigorous privacy and security guarantees has spurred a number of studies on machine learning via secure MPC~\cite{juvekar2018gazelle,kumar2020,mohassel2018,sadegh2017chameleon,cryptgpu,wagh2018securenn,wagh2021}. 
However, at present, adoption of secure MPC in machine learning is still relatively limited considering its wide-ranging potential.
One of the main obstacles to widespread adoption is that the complexity of secure MPC techniques puts them out of reach for most machine-learning researchers, who frequently lack in-depth knowledge of cryptographic techniques. 

To foster the adoption of secure MPC techniques in machine learning, we present \crypten{}: a flexible software framework that aims to make modern secure MPC techniques accessible to machine-learning researchers and developers without a background in cryptography.
Specifically, \crypten{} provides a comprehensive tensor-computation library in which all computations are performed via secure MPC.
\crypten{}'s API closely follows the API of the popular PyTorch framework for machine learning~\cite{paszke17,paszke2019}, which makes it easy to use for machine-learning practitioners.
For example, it provides automatic differentiation and a modular neural-network package.
\crypten{} assumes an \emph{semi-honest} threat model~\cite[\S2.3.2]{evans2018} and works for an arbitrary number of parties.
To make private training and inference efficient, \crypten{} off-loads computations to the GPU and uses high-performance communication libraries to implement interactions between parties.

The paper presents: (1) an overview of \crypten{}'s design principles; (2) a description of the design of \crypten{} and of the secure MPC protocols implemented; (3) a collection of benchmark experiments using \crypten{} to run private versions of state-of-the-art models for text classification, speech recognition, and image classification; and (4) a discussion of open problems and a roadmap for the further development of \crypten{}.
Altogether, the paper demonstrates that \crypten{}'s flexible, PyTorch-like API makes private inference and training of modern machine-learning models easy to implement and efficient.
For example, \crypten{} allows two parties to privately classify an image~\cite{dosovitskiy2021,he2016deep} in $2$-$3$ seconds, or to securely make phoneme predictions for 16kHz speech recordings~\cite{collobert2016} faster than real-time.
We hope that \crypten{}'s promising performance and ease-of-use will foster the adoption of secure MPC by the machine-learning community, and pave the way for a new generation of secure and private machine-learning systems.
\section{Related Work}
\label{sec:related_work}
\crypten{} is part of a large body of work that develops secure MPC protocols for machine learning; see Appendix~\ref{sec:framework_comparison}.
Most closely related to our work is CryptGPU~\cite{cryptgpu}, which implements an 2-out-of-3 replicated secret sharing protocol~\cite{araki2016,ito1989} \emph{on top of} \crypten{}. 
Like \crypten{}, CryptGPU provides security against \emph{semi-honest} corruption, but it is limited to the three-party setting. 
CryptGPU is one of several protocols optimized for the three-party setting. 
For example, Falcon~\cite{wagh2021} implements a \emph{maliciously secure} three-party MPC protocol, combining techniques from SecureNN~\cite{wagh2018securenn} and ABY3~\cite{mohassel2018}.
Falcon allows evaluation and training of convolutional networks such as AlexNet~\cite{krizhevsky2017imagenet} and VGG~\cite{simonyan2014}.
Other systems that work in this setting include Astra~\cite{chaudhari2019}, Blaze~\cite{patra2020}, and CrypTFlow~\cite{kumar2020}.

There also exists a family of two-party systems that, like \crypten{}, assume a {semi-honest} threat model.
These systems include Gazelle~\cite{juvekar2018gazelle}, Chameleon~\cite{sadegh2017chameleon}, EzPC~\cite{chandran2019}, MiniONN~\cite{liu2017minionn}, SecureML~\cite{mohassel2017}, PySyft~\cite{ryffel2018}, and Delphi~\cite{mishra2020delphi}.
XONN~\cite{sadegh2019} also works in the two-party setting but provides malicious security.
Compared to these systems, \crypten{} provides a more flexible machine-learning focused API\footnote{CrypTFlow~\cite{kumar2020} also provides such an API by integrating deeply with TensorFlow~\cite{tensorflow2015}, but unlike \crypten{}, it does not support PyTorch's eager execution model~\cite{paszke2019} or GPU support.} that supports reverse-mode automatic differentiation, implements a rich set of functions, and natively runs on GPUs.
Moreover, \crypten{} supports a wider range of use cases by working with an arbitrary number of parties, and make communication between parties efficient via communication primitives that were optimized for high-performance distributed computing.

\section{Design Principles}
\label{sec:design_principles}
In the development of \crypten{}, we adopted the following two main design principles:

\textbf{Machine-learning first API.} 
\crypten{} has a general purpose, machine-learning first API design. 
Most other secure MPC frameworks~\cite{mpc-sok} adopt an API that stays close to the underlying MPC protocols.
This hampers adoption of these frameworks in machine learning, for example, because they do not natively support tensor operations (but only scalar operations) and because they lack features that machine-learning researchers have come to expect, such as automatic differentiation.
Instead, \crypten{} implements the tensor-computation API of the popular PyTorch machine-learning framework~\cite{paszke17}, implements reverse-mode automatic differentiation, provides a modular neural-network package with corresponding learning routines, and supports GPU computations.
We aim to allow developers to transition code from PyTorch to \crypten{} by changing a single Python \texttt{import}.

\textbf{Eager execution.} 
\crypten{} adopts an imperative programming model.
This is different from existing MPC frameworks, which generally implement compilers for their own domain-specific languages~\cite{mpc-sok}.
While compiler approaches have potential performance benefits, they slow down the development cycle, make debugging harder, and prevent users from using arbitrary host-language constructs~\cite{agrawal2019tf}.
Instead, \crypten{} follows the recent trend in machine learning away from graph compilers~\cite{tensorflow2015} to frameworks that eagerly execute computations~\cite{agrawal2019tf,paszke2019}, providing a better developer experience.
Yet, \crypten{} is performant because it implements state-of-the-art secure MPC protocols (for settings with arbitrary number of parties), because it uses PyTorch's highly optimized tensor library for most computations, because computations can be off-loaded to the GPU, and because it uses communication libraries that were optimized for high-performance distributed computing.
%

\begin{figure}[t]
\centering
\includegraphics[width=0.83\linewidth]{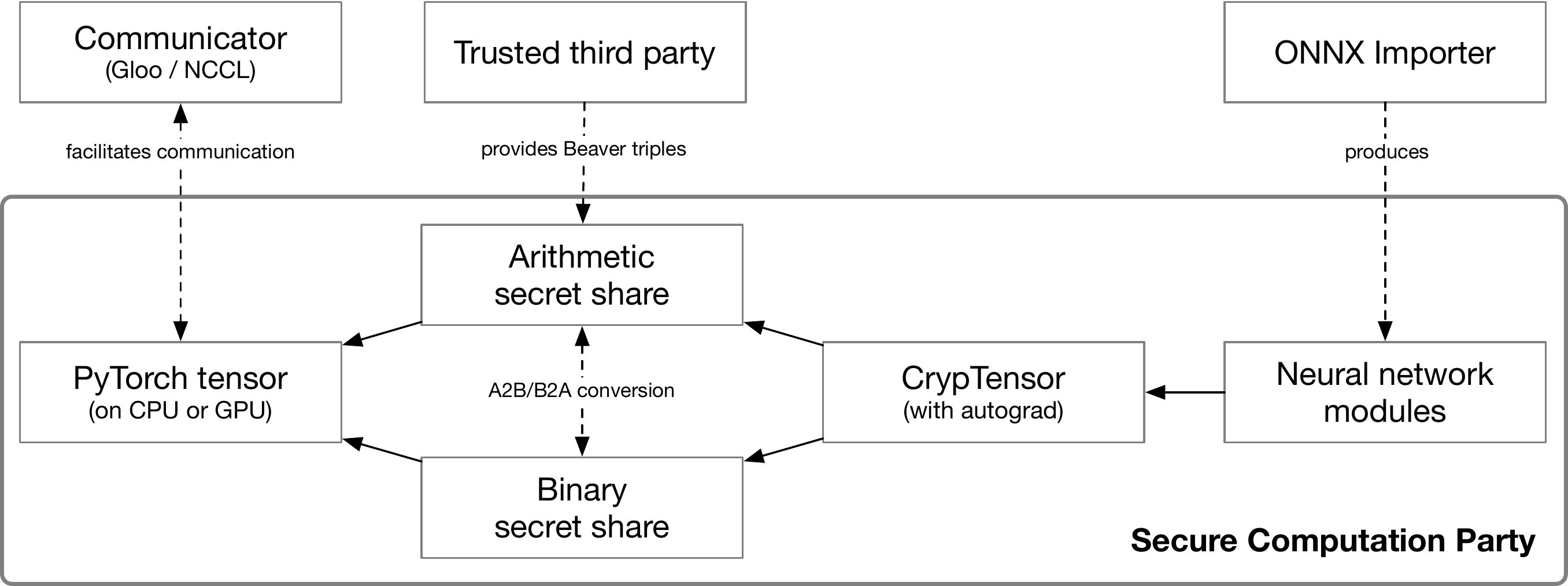}
\caption{High-level overview of the design of \crypten{}. See text in Section~\ref{sec:examples} for details.}
\label{fig:design}
\end{figure}

\section{Design Overview}
\label{sec:examples}

\begin{wrapfigure}{r}{0.5\textwidth}
\vspace{-4mm}
\begin{python}
import crypten, torch

# set up communication and sync random seeds:
crypten.init()  

# secret share tensor:
x = torch.tensor([1.0, 2.0, 3.0])
x_enc = crypten.cryptensor(x, src=0)  

# reveal secret shared tensor:
x_dec = x_enc.get_plain_text() 
assert torch.all_close(x_dec, x)

# add secret shared tensors:
y = torch.tensor([2.0, 3.0, 4.0])
y_enc = crypten.cryptensor(y, src=0)
xy_enc = x_enc + y_enc  
xy_dec = xy_enc.get_plain_text()
assert torch.all_close(xy_dec, x + y)
\end{python}
\caption{Example of secret-sharing tensors, revealing tensors, and private addition in \crypten{}.}
\label{fig:example1}
\vspace{-4mm}
\end{wrapfigure}

Figure~\ref{fig:design} gives an overview of \crypten{}'s design. 
Parties perform computations using efficient PyTorch tensor operations.
Because secure MPC computations are integer computations that are not natively supported on GPUs, \crypten{} maps between integer and floating-point computations on GPUs; see Section~\ref{sec:gpu_computations}.
The multi-party computations are implemented on arithmetic and binary secret shares~\cite{spdz,gmw}; see Section~\ref{sec:secret_sharing}.
Whereas many computations can be performed directly on arithmetic secret shares, others require conversion between arithmetic and binary secret shares (A2B) and back (B2A); see Section~\ref{sec:secure_computation}.
Some multi-party computations require interaction between parties via a \emph{communicator} that employs the high-performance communication primitives in Gloo~\cite{gloo} and NCCL~\cite{nccl}.
Some multi-party computations require Beaver triples~\cite{beaver1991}, which are supplied by a \emph{trusted third party} (TTP).\footnote{\crypten{} adopts a trusted third party for generating Beaver triples for efficiency reasons, but we are planning to add TTP-free solutions based on additive homomorphic encryption~\cite{paillier1999} or oblivious transfer~\cite{keller2016}.}

All secure computations are wrapped in a \texttt{CrypTensor} object that implements the PyTorch tensor API and that provides reverse-mode automatic differentiation (autograd) to enable gradient-based training of arbitrary (deep) learning models.
Figure~\ref{fig:example1} illustrates \texttt{CrypTensor} creation, \emph{i.e.}, how tensors are secret-shared and revealed, as well as a simple computation (addition).
Note that each party involved in the multi-party computation executes the same code.
Whenever communication between the parties is required (\emph{e.g.}, as part of private multiplications), the communication acts as a synchronization point between the parties.
The \texttt{crypten.init()} call is required once to establish the communication channel.
In the example, the input tensor for the creation of the arithmetic secret share is provided party \texttt{src=0}, which indicates the rank\footnote{\crypten{} relies on MPI primitives for communication: each party knows their rank and the world size.} of the party that supplies the data to be secret-shared (the other parties executing this code may provide \texttt{None} as input).

\begin{wrapfigure}{r}{0.5\textwidth}
\begin{python}
import crypten.optimizer as optimizer
import crypten.nn as nn

# create model, criterion, and optimizer:
model_enc = nn.Sequential(
    nn.Linear(sample_dim, hidden_dim),
    nn.ReLU(),
    nn.Linear(hidden_dim, num_classes),
).encrypt()
criterion = nn.CrossEntropyLoss()
optimizer = optimizer.SGD(
    model_enc.parameters(), lr=0.1, momentum=0.9,
)

# perform prediction on sample:
target_enc = crypten.cryptensor(target, src=0)
sample_enc = crypten.cryptensor(sample, src=0)
output_enc = model_enc(sample_enc)

# perform backward pass and update parameters:
model_enc.zero_grad()
loss_enc = criterion(output_enc, target_enc)
loss_enc.backward()
optimizer.step()
\end{python}
\caption{Example using neural networks and automatic differentiation in \crypten{}.}
\label{fig:example2}
\vspace{4mm}
\begin{python}
import torchvision.datasets as datasets
import torchvision.models as models
import torchvision.transforms as transforms

# download and set up ImageNet dataset:
transform = transforms.ToTensor()
dataset = datasets.ImageNet(
    imagenet_folder, transform=transform,
)

# secret share pre-trained ResNet-18 on GPU:
model = models.resnet18(pretrained=True)
model_enc = crypten.nn.from_pytorch(
    model, dataset[0],
).encrypt().cuda()

# perform inference on secret-shared images:
for image in dataset:
    image_enc = crypten.cryptensor(image).cuda()
    output_enc = model_enc(image_enc)
    output = output_enc.get_plain_text()
\end{python}
\caption{Private inference on secret-shared images using a secret-shared ResNet-18 model on GPU.}
\label{fig:example3}
\vspace{-22mm}
\end{wrapfigure}

To enable deep-learning use cases, \crypten{} allows implementing neural networks following PyTorch's API.
Figure~\ref{fig:example2} shows how to create and encrypt neural networks and how to use automatic differentiation in \crypten{}.
The example assumes that some training {\ttmlarge sample} and the associated {\ttmlarge target} label are provided by the party with rank 0 (note the value of {\ttmlarge src}).
As illustrated by the example, \crypten{}'s API closely follows that of PyTorch.
Indeed, it is possible to write a single training loop that can be used to train models using \crypten{} or PyTorch without code changes.
This makes it easy to adapt PyTorch code to use secure MPC for its computations, and it also makes debugging easier.
The appendix presents a table listing all tensor functions that \texttt{CrypTensor} implements.

To enable interoperability with existing machine-learning platforms, neural networks can be imported into \crypten{} via ONNX.
Figure~\ref{fig:example3} shows how a PyTorch model is imported into \crypten{}.
The example illustrates how \crypten{} makes private inference with a ResNet-18 easy.
The example in the figure also demonstrates \crypten{}'s GPU support.
One caveat is that all parties must use the same type of device (\emph{i.e.}, CPU or GPU) for computations.
\section{Secure Computations}
\label{sec:protocols_highlevel}
To facilitate secure computations, \crypten{} implements arithmetic secret sharing~\cite{spdz,damgrd2019new} and binary secret sharing~\cite{gmw}, as well as conversions between these two types of sharing~\cite{demmler2015aby}.
Arithmetic secret sharing is particularly well-suited for operations that are common in modern machine-learning models, such as matrix multiplications and convolutions.
Binary secret sharing is required for evaluating certain other common functions, such as rectified linear units.
We provide a high-level overview of \crypten{}'s secure computation protocol here; a detailed description is presented in the appendix.

\subsection{Secret Sharing}
\label{sec:secret_sharing}

\noindent\textbf{Arithmetic secret sharing} shares a scalar value $x \in \thefield$, where $\thefield$ denotes a ring with $Q$ elements, across parties $p \in \mathcal{P}$.
We denote the sharing of $x$ by $[x] = \{[x]_p\}_{p \in \mathcal P}$, where $[x]_p \in \thefield$ indicates party $p$'s share of $x$. 
The shares are constructed such that their sum reconstructs the original value $x$, that is, $x = \sum_{p \in \mathcal{P}} [x]_p \mod Q$.
To share a value $x$, the parties generate a pseudorandom zero-share \cite{cramer2005} with $| \mathcal P |$ random numbers that sum to $0$.
The party that possesses the value $x$ adds $x$ to their share and discards $x$.
We use a fixed-point encoding to obtain $x$ from a floating-point value, $x_R$. 
To do so, we multiply $x_R$ with a large scaling factor $B$ and round to the nearest integer: $x = \lfloor B x_R \rceil$, where $B = 2^L$ for some precision of $L$ bits. 
To decode a value, $x$, we compute $x_R \approx \nicefrac{x}{B}$.

\noindent\textbf{Binary secret sharing} is a special case of arithmetic secret sharing that operates within the binary field $\mathbb{Z}/2\mathbb{Z}$.
A binary secret share, $\langle x \rangle$, of a value $x$ is formed by arithmetic secret shares of the bits of $x$, setting $Q\!=\!2$. Each party $p \in \mathcal P$ holds a share, $\langle x\rangle_p$, such that $x = \bigoplus_{p \in \mathcal P} \langle x \rangle_p $ is satisfied.

\noindent\textbf{Conversion from $\boldsymbol{[x]}$ to $\boldsymbol{\langle x\rangle}$} is implemented by having the parties create a binary secret share of their $[x]_p$ shares, and summing the resulting binary shares.
Specifically, the parties create a binary secret share, $\langle [x]_p \rangle$, of all the bits in $[x]_p$. 
Subsequently, the parties compute $\langle x \rangle = \sum_{p\in \mathcal P} \langle[x]_p\rangle$ using a carry-lookahead adder in $\log_2 (|\mathcal P|)\log_2(Q)$ communication rounds~\cite{catrina2010improved, damgrd2005unconditionally}.

\noindent\textbf{Conversion from $\boldsymbol{\langle x\rangle}$ to $\boldsymbol{[x]}$} is achieved by computing $[x] = \sum_{b=1}^B 2^b \left[\langle x\rangle^{(b)}\right]$, where $\langle x\rangle^{(b)}$ denotes the $b$-th bit of the binary share $\langle x\rangle$ and $B$ is the total number of bits in the shared secret, $\langle x\rangle$.
To create an arithmetic share of a bit, the parties use secret shares, $\left([r^{(b)}], \langle r^{(b)} \rangle\right)$, of random bits $r^{(b)}$.
The random bits are provided by the TTP, but we plan to add an implementation that generates them off-line via oblivious transfer~\cite{keller2016}.
The parties use $\langle r^{(b)} \rangle$ to mask $\langle x\rangle^{(b)}$ and reveal the resulting masked bit $z^{(b)}$.
Subsequently, they compute $\left[\langle x\rangle^{(b)}\right] = \left[r^{(b)}\right] + z^{(b)} - 2 \left[r^{(b)}\right] z^{(b)}$.

\subsection{Secure Computation}
\label{sec:secure_computation}
Arithmetic and binary secret shares have homomorphic properties that can be used to implement secure computations.
All computations in \crypten{} are based on private addition and multiplication.

\noindent\textbf{Private addition} of two arithmetically secret shared values, $[z] = [x] + [y]$, is implemented by having each party $p$ sum their shares of $[x]$ and $[y]$: each party $p \in \mathcal{P}$ computes $[z]_p = [x]_p + [y]_p$. 

\noindent\textbf{Private multiplication} is implemented using random Beaver triples \cite{beaver1991}, $([a], [b], [c])$ with $c \!=\!
ab$, that are provided by the TTP. 
The parties compute $[\epsilon] = [x] - [a]$ and $[\delta] = [y] - [b]$, and decrypt $\epsilon$ and $\delta$ without information leakage due to the masking. 
They compute the result $[x][y] = [c] + \epsilon[b] + [a]\delta + \epsilon\delta$, using trivial implementations of addition and multiplication of secret shares with public values.

\noindent\textbf{Linear functions} are trivially implemented as combinations of private addition and multiplication.
This allows \crypten{} to compute dot products, outer products, matrix products, and convolutions.

\noindent\textbf{Non-linear functions} are implemented using standard approximations that only require private addition and multiplication.
Specifically, \crypten{} evaluates exponentials using a limit approximation, logarithms using Householder iterations~\cite{householder1970}, and reciprocals using Newton-Rhapson iterations.
This allows \crypten{} to implement functions that are commonly used in machine-learning models, including the sigmoid, softmax, and logistic-loss functions, as well as their gradients.

\noindent\textbf{Comparators} are implemented using a function that evaluates $[z<0]$ by: (1) converting $[z]$ to a binary secret-share $\langle z \rangle$; (2) computing its sign bit, $\langle b \rangle = \langle z \rangle >\!> (L - 1)$; and (3) converting
the resulting bit to an arithmetic sharing $[b]$.
This function allows \crypten{} to implement arbitrary comparators.
For example, it evaluates $[x < y]$ by computing $[z] = [x] - [y]$ and evaluating $[z<0]$.
Similarly, \crypten{} can evaluate: (1) the sign function via $\operatorname{sign}([x]) = 2[x > 0] - 1$; (2) the absolute value function via $|[x]| = [x]\operatorname{sign}([x])$; and (3) rectified linear units via $\operatorname{ReLU}([x]) = [x][x > 0]$.
\crypten{} also supports multiplexing; to do so, it evaluates $[c~?~x : y] = [c][x] + (1 - [c])[y]$.

\begin{lemma}\label{thm:security}
The \crypten{} secure-computation protocol is secure against information leakage against any static passive adversary corrupting up to $|\mathcal{P}|-1$ of the $|\mathcal{P}|$ parties involved in the computation.
\end{lemma}
The proof of this lemma follows trivially from~\cite{sharemind2008,canetti2001,damgrd2005unconditionally,demmler2015aby}, and is given in the appendix.
We adopt a protocol that provides security under a \emph{semi-honest} threat model because it enables a wide range of use cases of secure machine learning, whilst being more efficient than maliciously secure protocols.

\subsection{Off-loading Computations to the GPU}
\label{sec:gpu_computations}
Hardware acceleration via GPUs is a critical component for training and inference in modern machine-learning models.  
Akin to frameworks such as PyTorch \cite{paszke2019} and TensorFlow \cite{tensorflow2015}, \crypten{} can off-load computations to the GPU. 
On the GPU, it uses highly-optimized implementations for a range of functions that are provided by CUDA libraries such as cuBLAS~\cite{cublas} and cuDNN~\cite{cudnn}. 

Unfortunately, these libraries are designed for computations on floating-point numbers and do not support the integer types required to perform computations on $L$-bit fixed-point numbers. 
Akin to~\cite{cryptgpu}, we circumvent this problem by observing that for all integers $a, b \in \mathbb{Z} \cap [-2^{26}, 2^{26}]$, we can compute the product $ab$ using 64-bit floating-point representations and still recover the correct value over the integers. 
Specifically, \crypten{} splits each 64-bit variable into four components, $a = a_0 + 2^{16}a_1 + 2^{32}a_2 + 2^{48}a_3$, where each $a_i$ represents a 16-bit integer component. 
We compute a product $ab$ of 64-bit integers by summing 10 pairwise products of their 16-bit components. 
The pairwise products of the 16-bit components are computed in parallel using highly optimized floating-point CUDA kernels. 
The same approach is used for matrix multiplications and convolutions. 
\crypten{} further optimizes this approach by splitting into only 3 components of 22-bits each when possible, which reduces the number of pairwise products required to 6 (see \cite[Remark II.1]{cryptgpu}).
\section{Benchmarks}
\label{sec:benchmarks}
To measure the performance of \crypten{}, we performed experiments on three tasks: (1) text classification using a linear model that learns word embeddings; (2) speech recognition using the Wav2Letter model \cite{collobert2016}; and (3) image classification using residual networks \cite{he2016deep} and vision transformers \cite{dosovitskiy2021}.
Because of space constraints, we focus on private inference using a secret-shared model on secret-shared data here, but our benchmark results with private training are very similar.

We performed benchmark experiments on a proprietary cluster, testing inference on both CPUs (Intel Skylake 18-core 1.6GHz) and GPUs (nVidia P100).
We set the number of OpenMP threads to 1 in all benchmarks.
All experiments were performed with the parties running in separate processes on a single machine.
For GPU experiments, each party was assigned its own GPU.
Although this setup is faster than a scenario in which each party operates its own machine,\footnote{Communication between GPUs in two machines connected via InfiniBand has approximately $20\times$ lower throughput than communication between two GPUs in the same machine via NVLink (25GB/s versus 600GB/s).} we believe our benchmark results provide a good sense of \crypten{}'s performance.
We average computation times over 30 batches, excluding the computation on the first batch as that computation may include CuDNN benchmarking.
Code reproducing the results of our experiments is available on \url{https://crypten.ai}.

In our benchmarks, we focus on comparing (ciphertext) \crypten{} computation with (plaintext) PyTorch computation.
We refer the reader to~\cite{haralampieva2020,cryptgpu} for benchmarks that compare \crypten{} to other secure MPC frameworks.
Specifically, \cite{haralampieva2020} finds \crypten{} is $11$-$18\times$ faster than PySyft~\cite{ryffel2018} and approximately $3\times$ faster than TF-Trusted~\cite{tftrusted} in MNIST classification~\cite{lecun1998} on CPU.

\begin{figure}[t]
\centering
\includegraphics[width=0.32\linewidth]{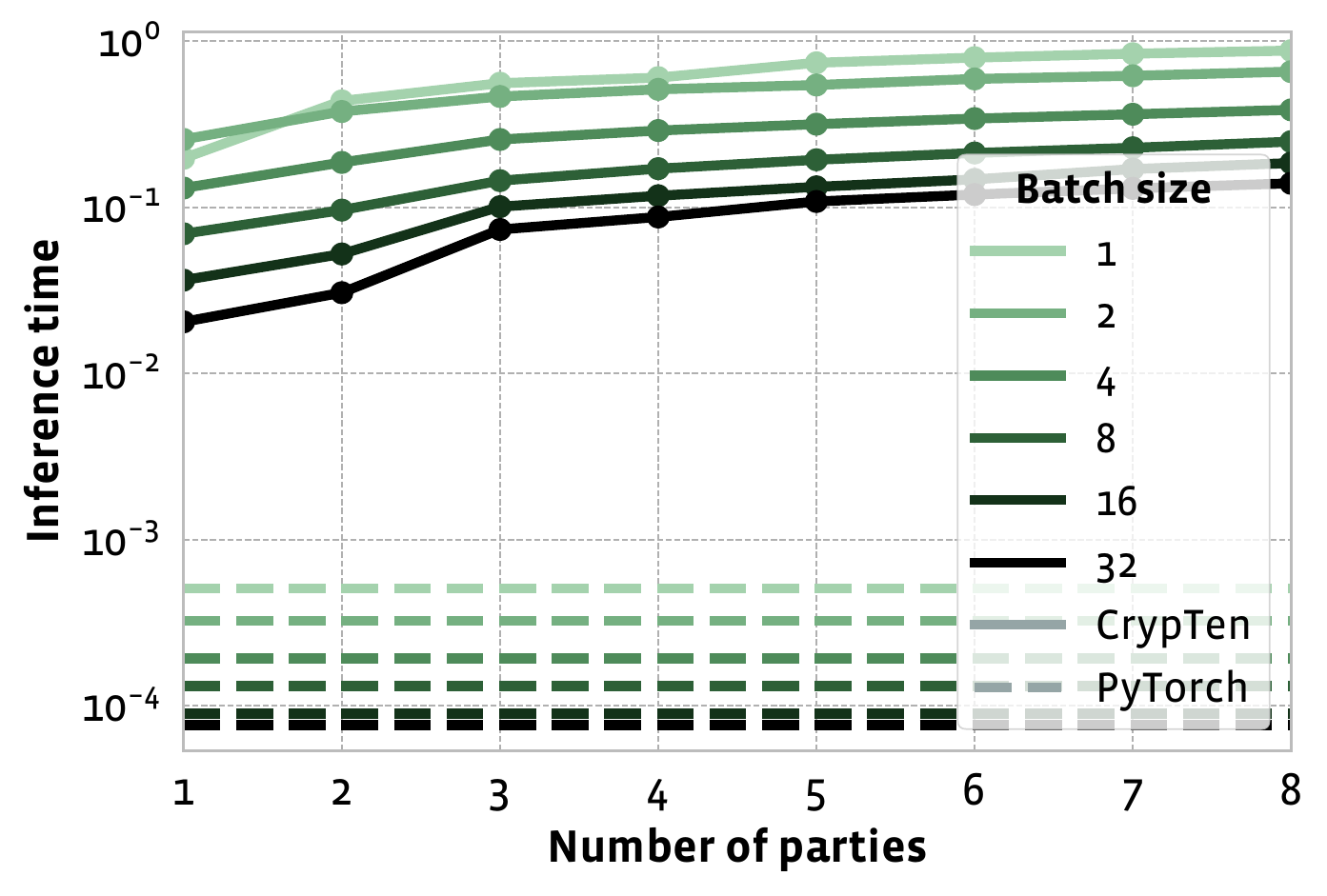}
\includegraphics[width=0.32\linewidth]{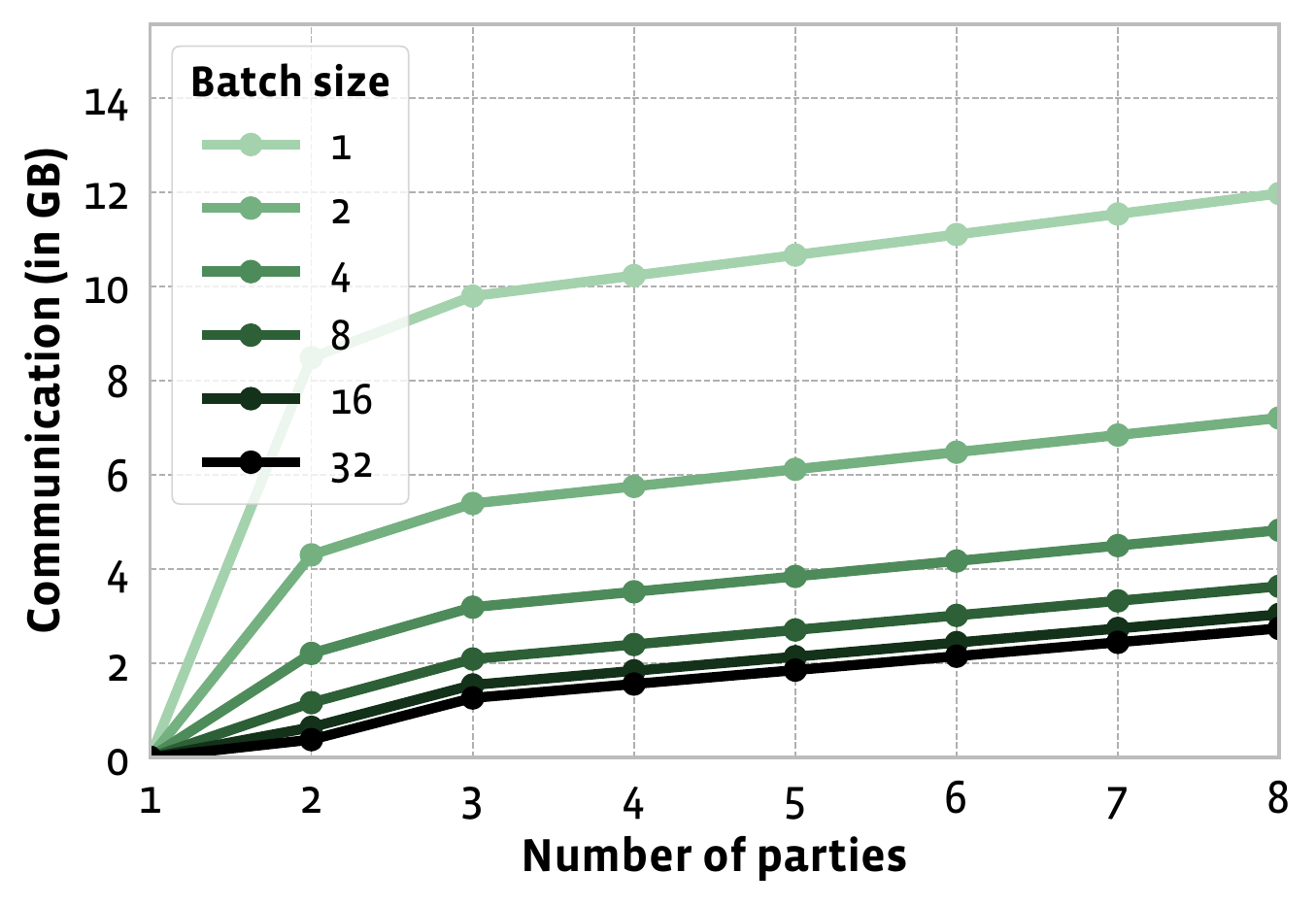}
\includegraphics[width=0.32\linewidth]{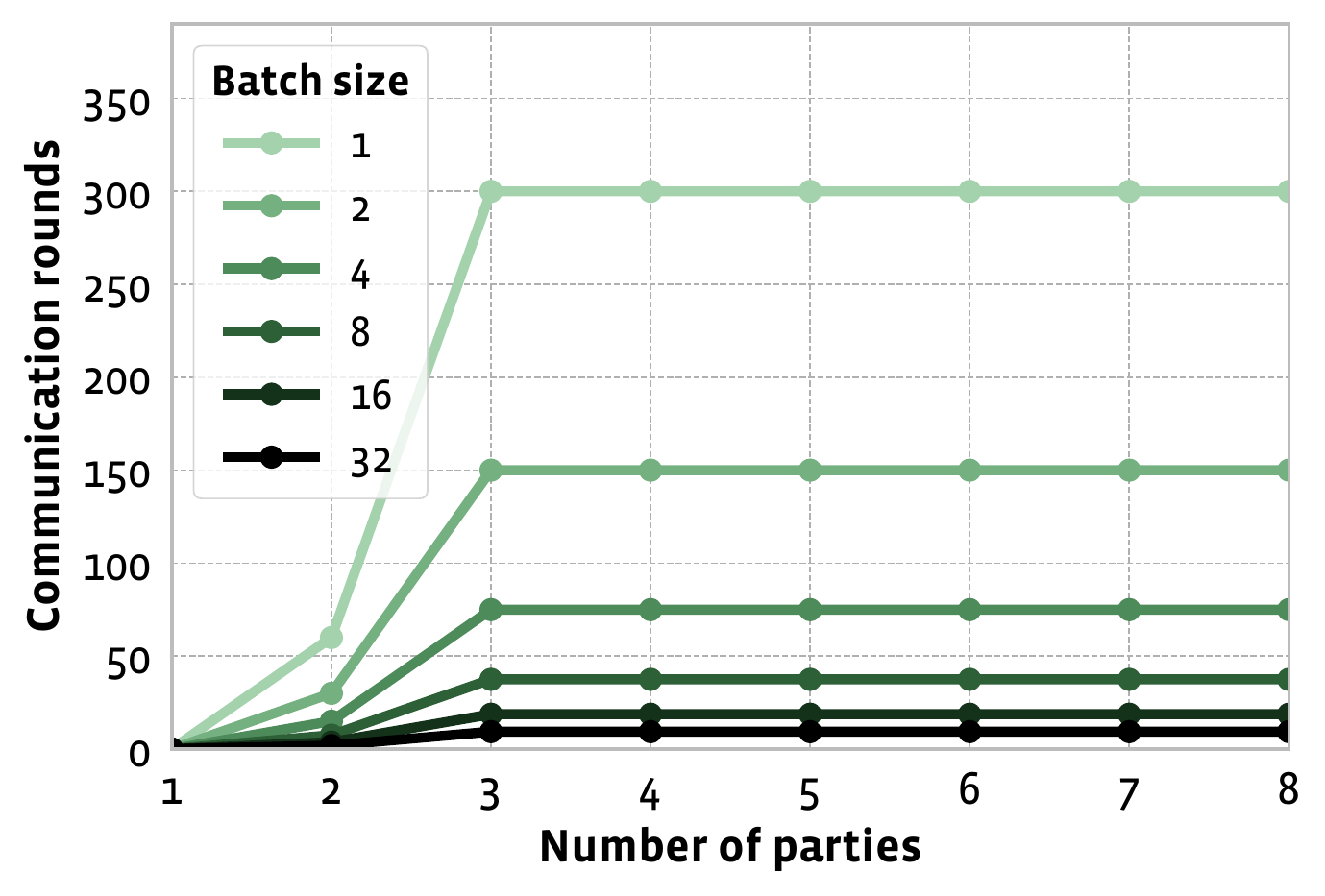}
\caption{Benchmarks for inference with text-sentiment classification model on GPUs in \crypten{} and PyTorch. \textbf{Left:} Average wall-clock time per sample (in seconds). \textbf{Middle:} Number of bytes communicated per sample, per party (in GB). \textbf{Right:} Number of communication rounds per sample.}
\label{fig:sentiment}
\end{figure}

\subsection{Text Classification}
We performed text-sentiment classification experiments on the Yelp review dataset~\cite{yelpreview} using a model that consists of a linear layer operating on word embeddings. 
The embedding layer contains $32$-dimensional embeddings of $519,820$ words, and the linear layer produces a binary output indicating the sentiment of the review.
We evaluated the model on GPUs, varying the batch size and the number of parties participating.
The normalized mean squared error ($\nicefrac{\lVert \bx - \by \rVert^2}{\lVert \bx \rVert^2}$) between the output of the \crypten{} model and that of its PyTorch counterpart was smaller than $4 \cdot 10^{-4}$ in all experiments.

Figure~\ref{fig:sentiment} presents the results of our experiments.
The figure shows inference time \emph{per sample} (in seconds) as a function of the number of parties involved in the computation for varying batch sizes (left); the amount of communication required per sample, \emph{per party} (in GB); and the number of communication rounds required per sample.
We include results in which the number of parties is $1$: herein, we run the \crypten{} protocol but involve no other parties, which implies that the single party is running the protocol on unencrypted data.
One-party results allow us to bisect different sources of computational overhead: specifically, they separate overhead due to communication from overhead due to fixed-point encoding, function approximations, and (lack of) sparse-matrix operations.

The results in Figure~\ref{fig:sentiment} show that \crypten{} is about $2.5$--$3$ orders of magnitude slower than PyTorch in text-sentiment classification, depending on the number of parties involved.
Most computational overhead is the word embedding layer: whereas PyTorch can evaluate this layer efficiently via a sparse matrix multiplication, \crypten{} cannot do sparse lookups as they would reveal information on the encrypted input.
Instead, \crypten{} performs a full matrix multiplication between the word-count vector and the embedding matrix.
Yet, text sentiment predictions are quite fast in \crypten{}: inference takes only $0.03$ seconds per sample in the two-party setting with a batch size of $32$.

The results also show that increasing the batch size is an effective way to reduce inference time and communication per sample.
The number of communication rounds is independent of the batch size, which means communication rounds can be amortized by using larger batch sizes.
The number of bytes communicated is partly amortized as well because the size of weight tensors (\emph{e.g.}, in linear layers) does not depend on batch size. 
The results also show that whereas the number of communication rounds increases when moving from two-party to three-party computation, it remains constant afterwards.
The larger number of communication rounds for three-party computation stems from the public division protocol, which requires additional communication rounds when more than two parties are involved to prevent wrap-around errors (see the appendix for details). 

\begin{figure}[t]
\centering
\includegraphics[width=0.32\linewidth]{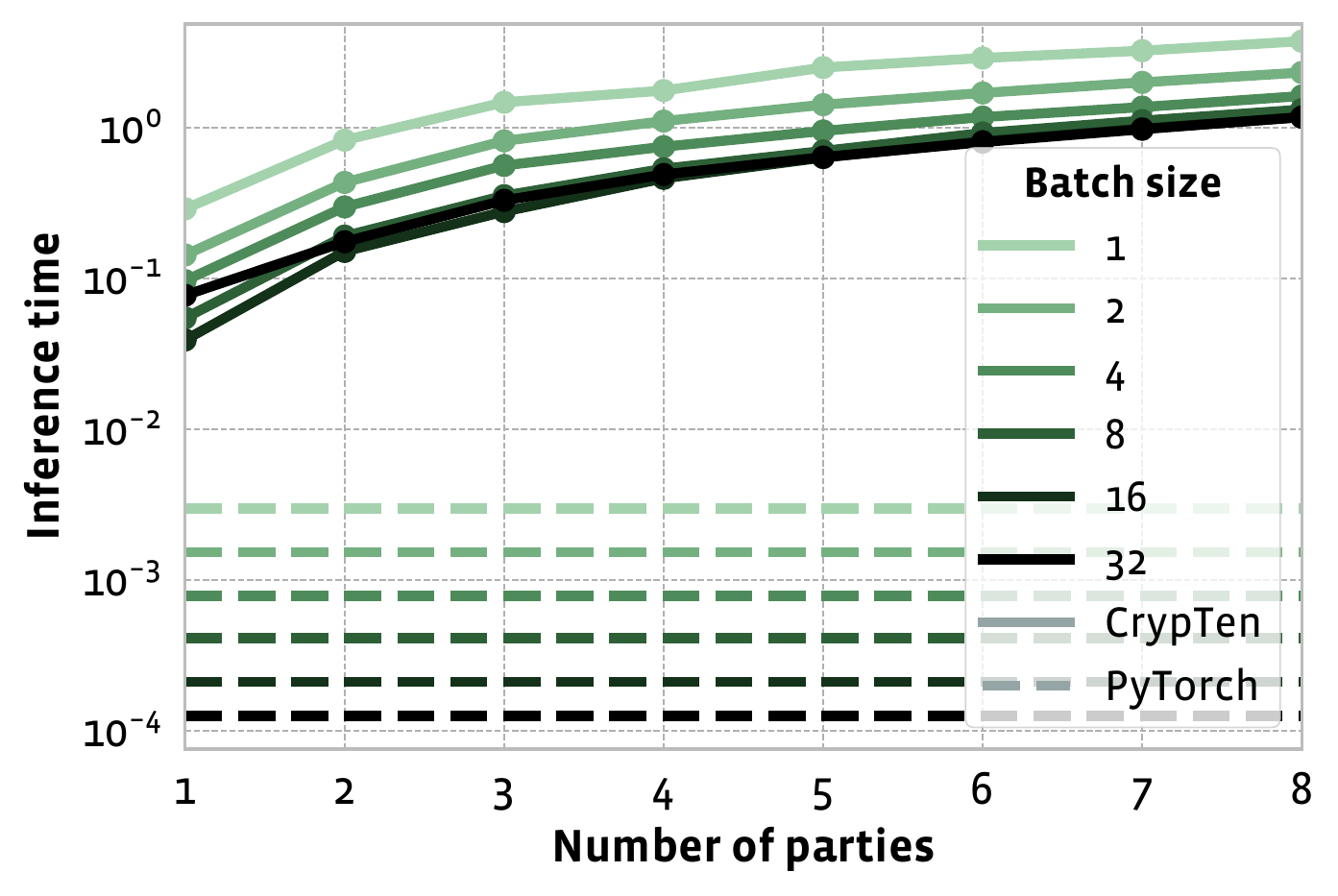}
\includegraphics[width=0.32\linewidth]{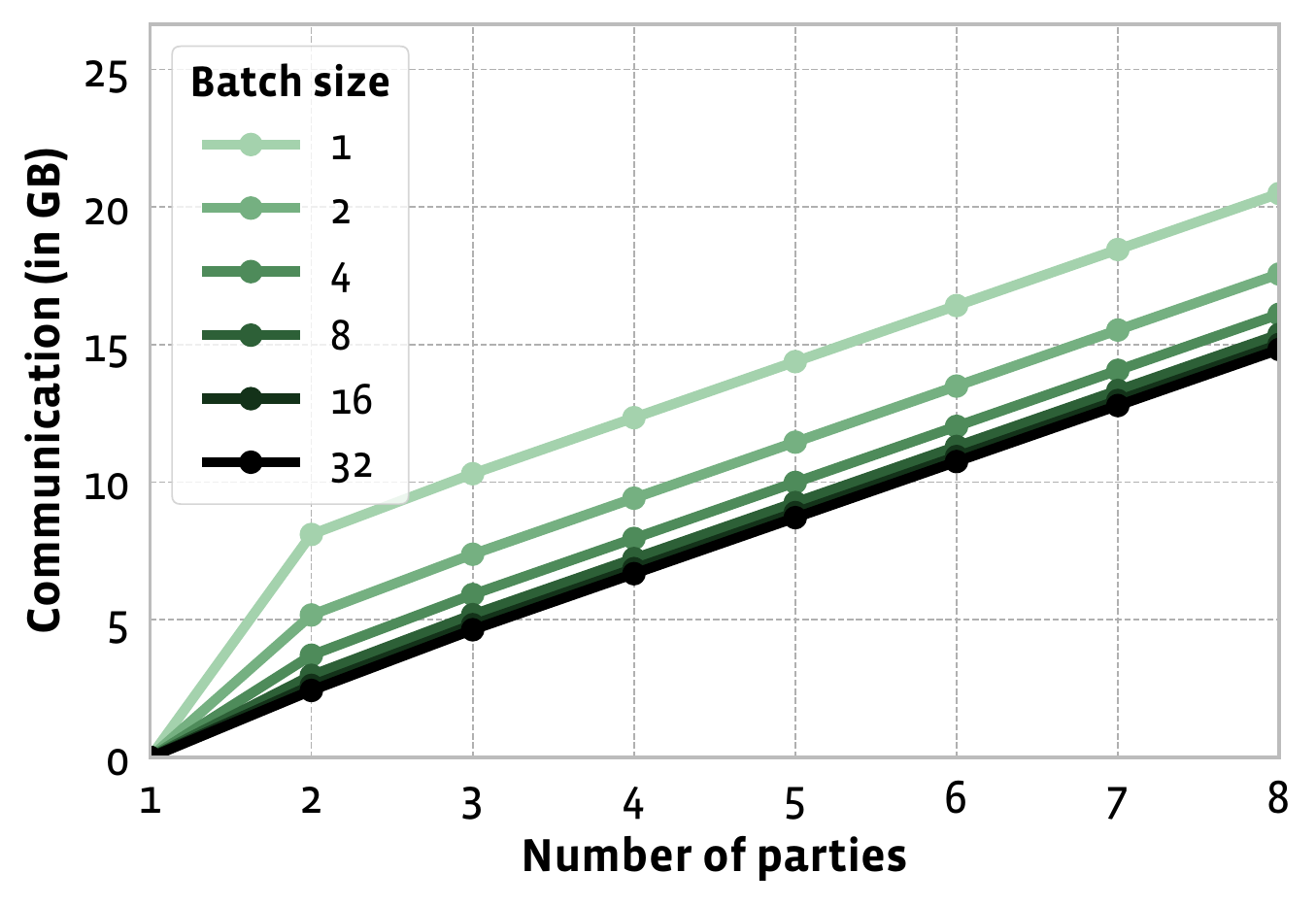}
\includegraphics[width=0.32\linewidth]{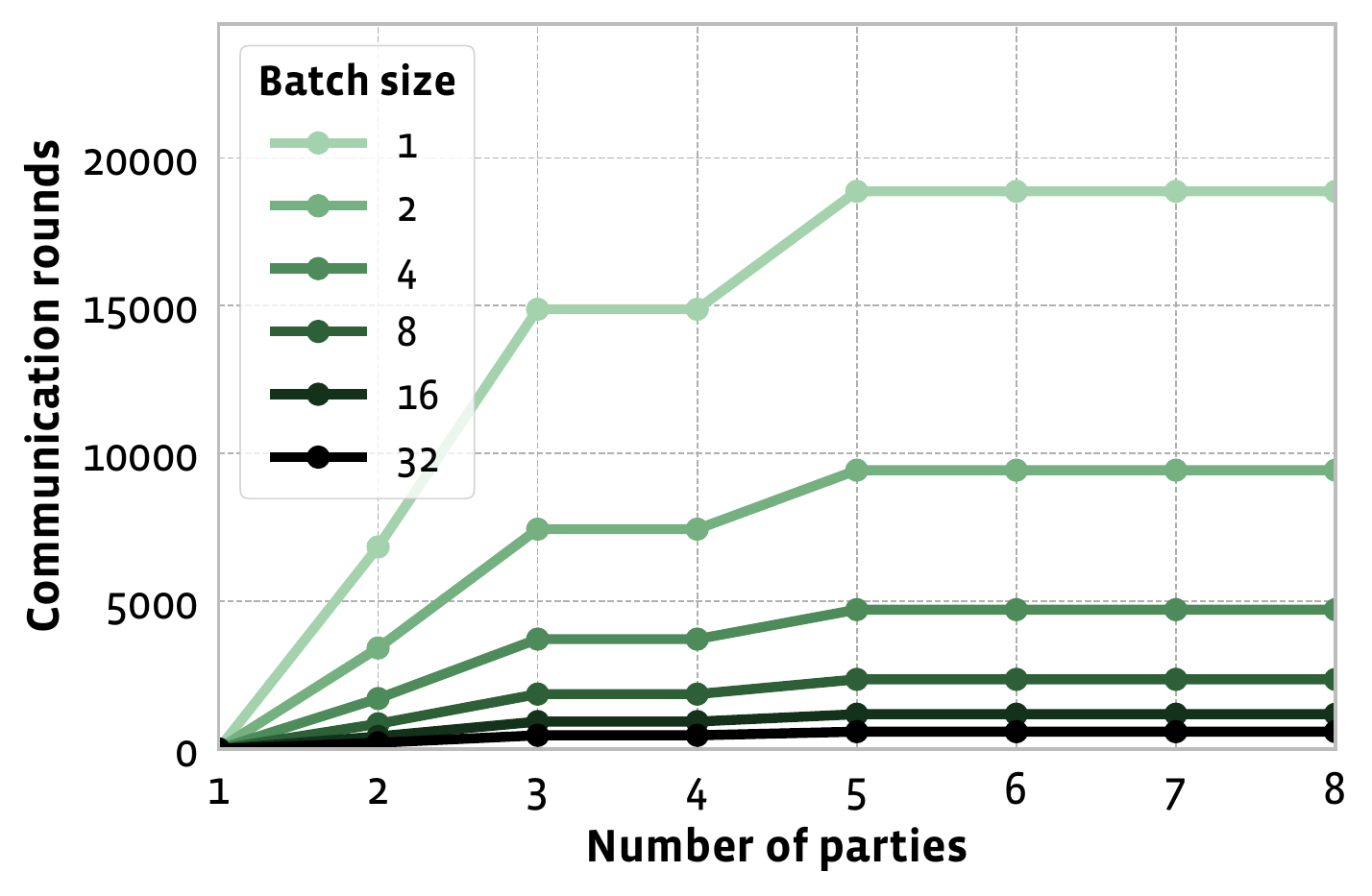}
\caption{Benchmarks for inference with Wav2Letter model on GPUs in \crypten{} and PyTorch. \textbf{Left:} Average wall-clock time per sample (in seconds). \textbf{Middle:} Number of bytes communicated per sample, per party (in GB). \textbf{Right:} Number of communication rounds per sample.}
\label{fig:wav2letter}
\end{figure}

\begin{figure}[t]
\centering
\begin{minipage}{.32\textwidth}
    \includegraphics[width=\linewidth]{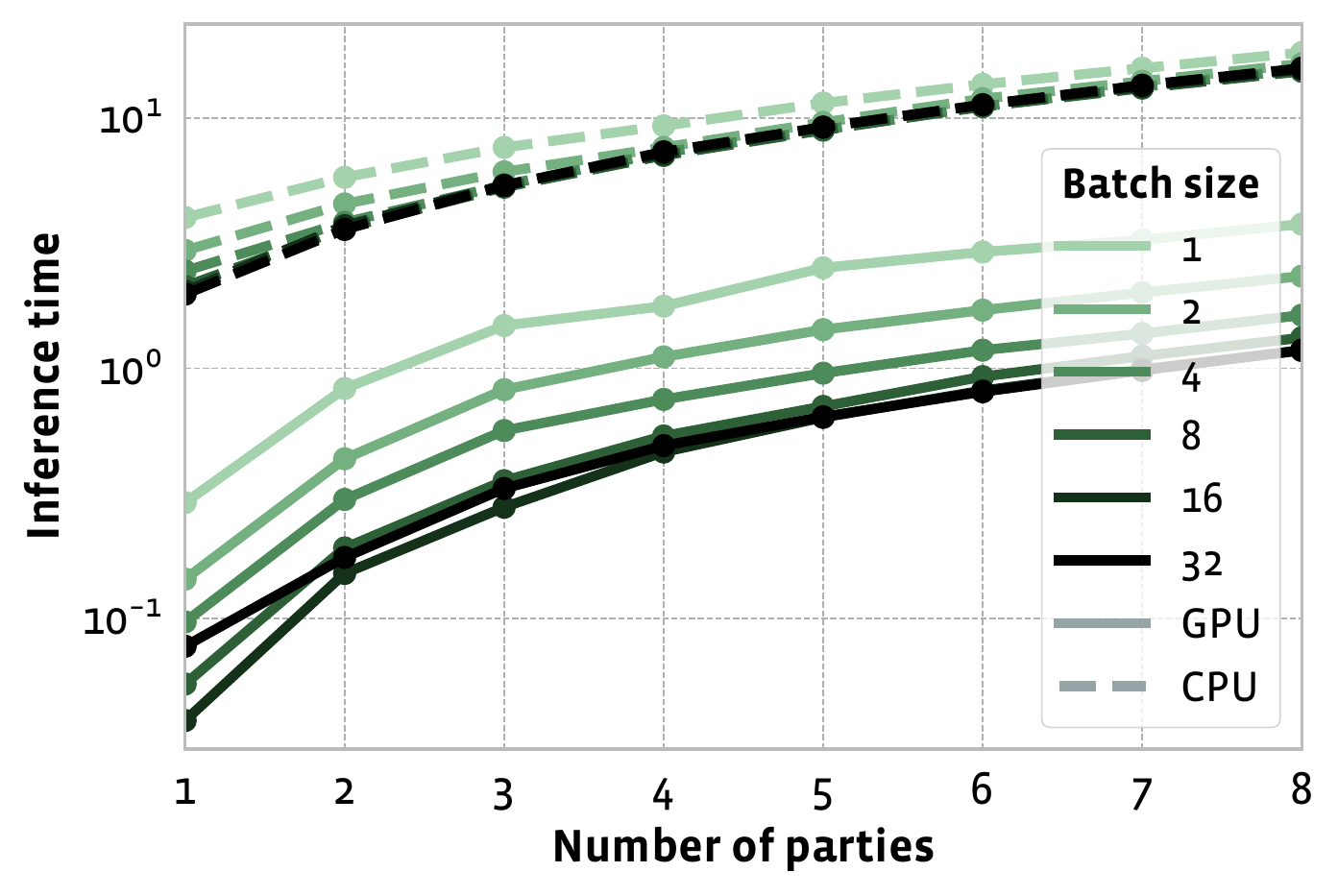}
    \caption{Wall-clock time per sample (in sec.) for Wav2Letter inference on CPUs and GPUs.}
    \label{fig:wav2letter_devices}
\end{minipage}
\hspace{.02\textwidth}
\begin{minipage}{.64\textwidth}
  \includegraphics[width=0.48\linewidth]{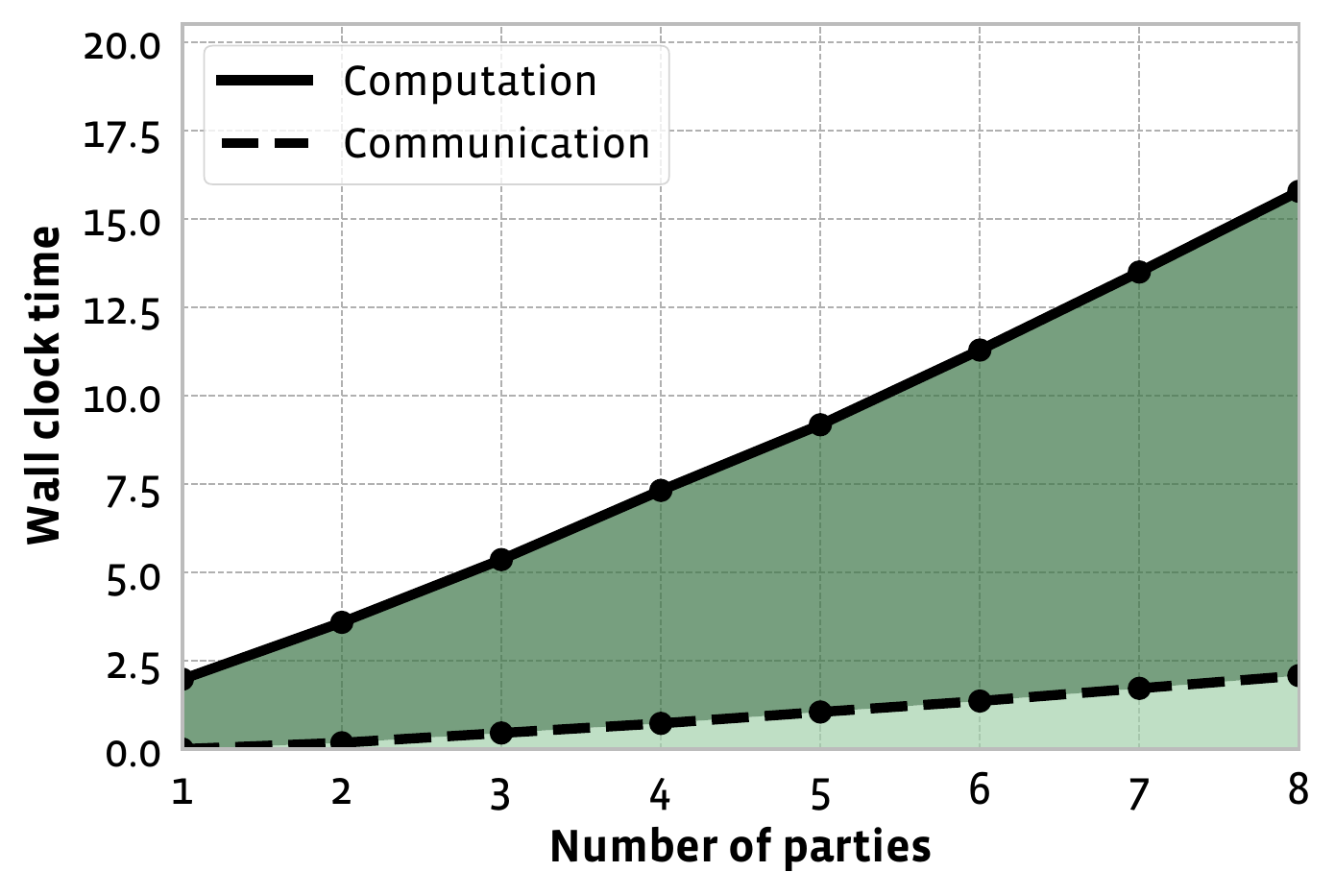}
  \includegraphics[width=0.48\linewidth]{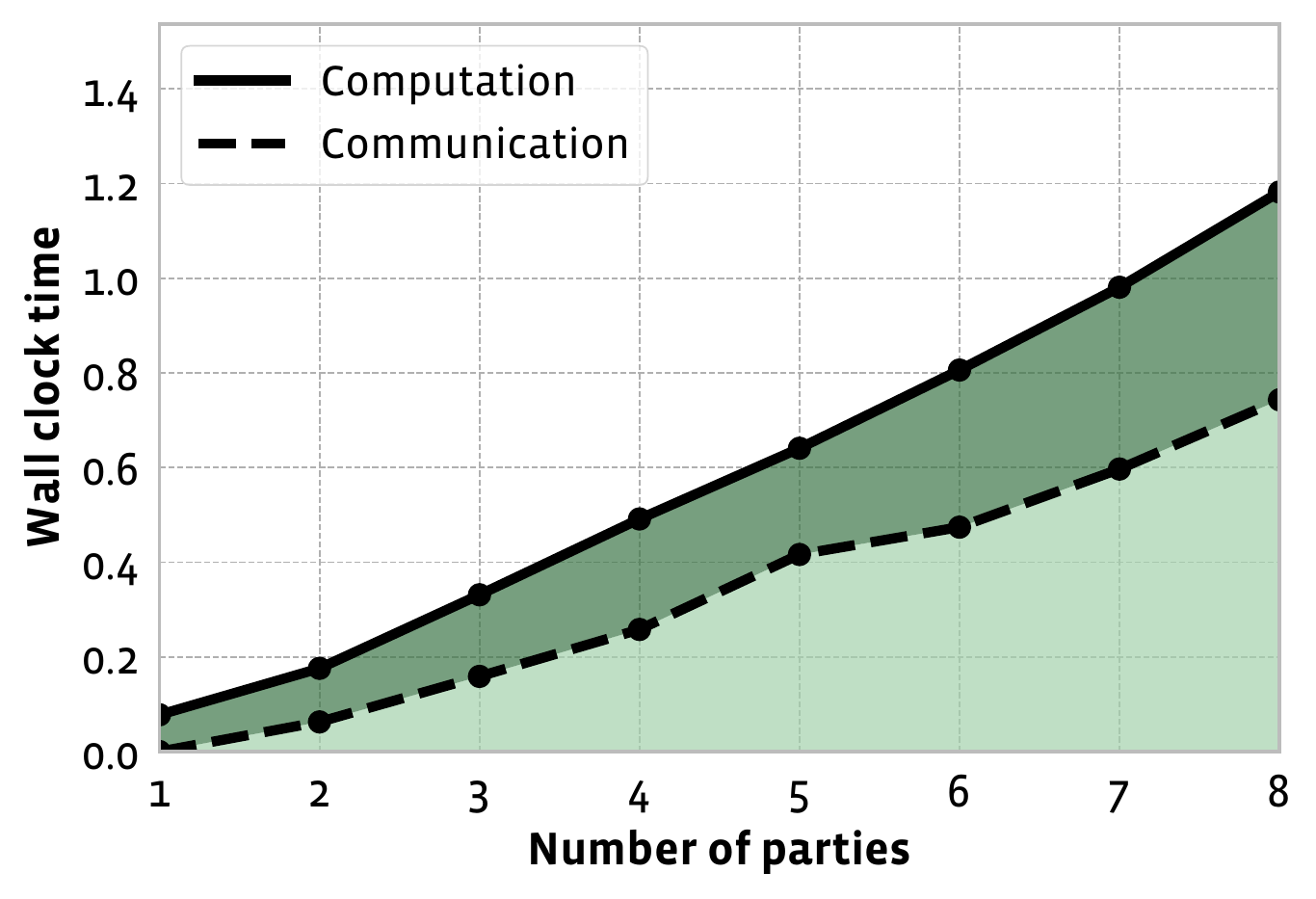}
  \caption{Average wall-clock time per sample (in seconds) for communication and computation during inference with Wav2Letter model on CPU (\textbf{left}) and GPU (\textbf{right}).}
  \label{fig:wav2letter_communication}
\end{minipage}
\end{figure}

\subsection{Speech Recognition}
We performed speech-recognition experiments using Wav2Letter~\cite{collobert2016} on the LibriSpeech dataset~\cite{panayotov2015}. 
The LibriSpeech dataset contains $16$ kHz audio clips represented as a waveform ($16,000$ samples per second).
Because the audio clips vary in length, we clip all of them to $1$ second for the benchmark.
Wav2Letter is a network with 13 convolutional layers using rectified linear unit (ReLU;~\cite{nair2010}) activations.\footnote{We used the reference implementation of Wav2Letter in \texttt{torchaudio}.}
The network operates directly on the waveform input, predicting one of 29 labels (26 letters plus 3 special characters).
The first two layers use a filter size of $250$ (with stride $160$) and $48$ (stride $2$).
The next seven layers use filter size $7$, followed by two layers with filter size $32$ and $1$ (all with stride $1$).
All layers except the last two have $250$ channels. 
The last two layers have $2,000$ channels.

The results in Figure~\ref{fig:wav2letter} show that \crypten{} is about $2.5$--$3$ orders of magnitude slower than PyTorch depending on the number of parties involved.
For Wav2Letter, the overhead is largely due to the ReLU layers in the network: evaluating a ReLU function requires a comparison, which involves a conversion between arithmetic and binary secret sharing and back (see the appendix).
The number of communication rounds increases when the number of parties grows beyond $4$: \crypten{} uses a tree reduction for the summation in the comparator protocol, which implies that the number of communication rounds grows whenever the number of parties increases from $2^k$ to $2^{k+1}$.

Figure~\ref{fig:wav2letter_devices} also presents results comparing Wav2Letter inference time between CPUs and GPUs.
The results in the figure show that \crypten{} is $1$-$2$ orders of magnitude faster on GPUs than on CPUs.
In real-world settings, this speedup can make the difference between a secure MPC use case being practical or not.
Figure~\ref{fig:wav2letter_communication} shows how much wall-clock time is spent on communication and computation, respectively, when performing inference with Wav2Letter (using batch size 32).
The results suggest that, whereas multi-party evaluation is compute-bound on CPU, it is communication-bound on GPU.
On GPUs, $63$\% of the time is spent on communication in eight-party computation.

\begin{figure}[t]
\centering
\includegraphics[width=0.32\linewidth]{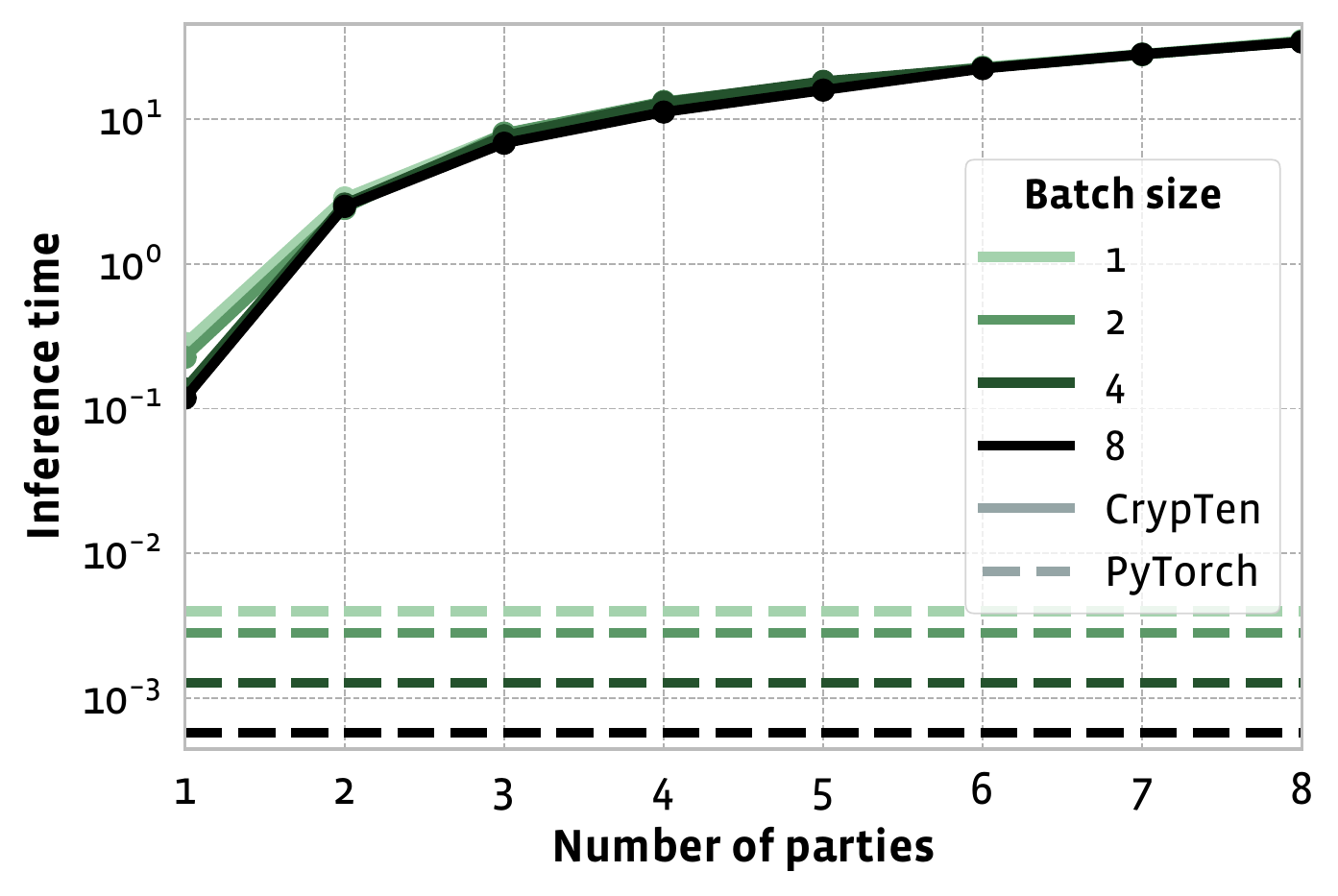}
\includegraphics[width=0.32\linewidth]{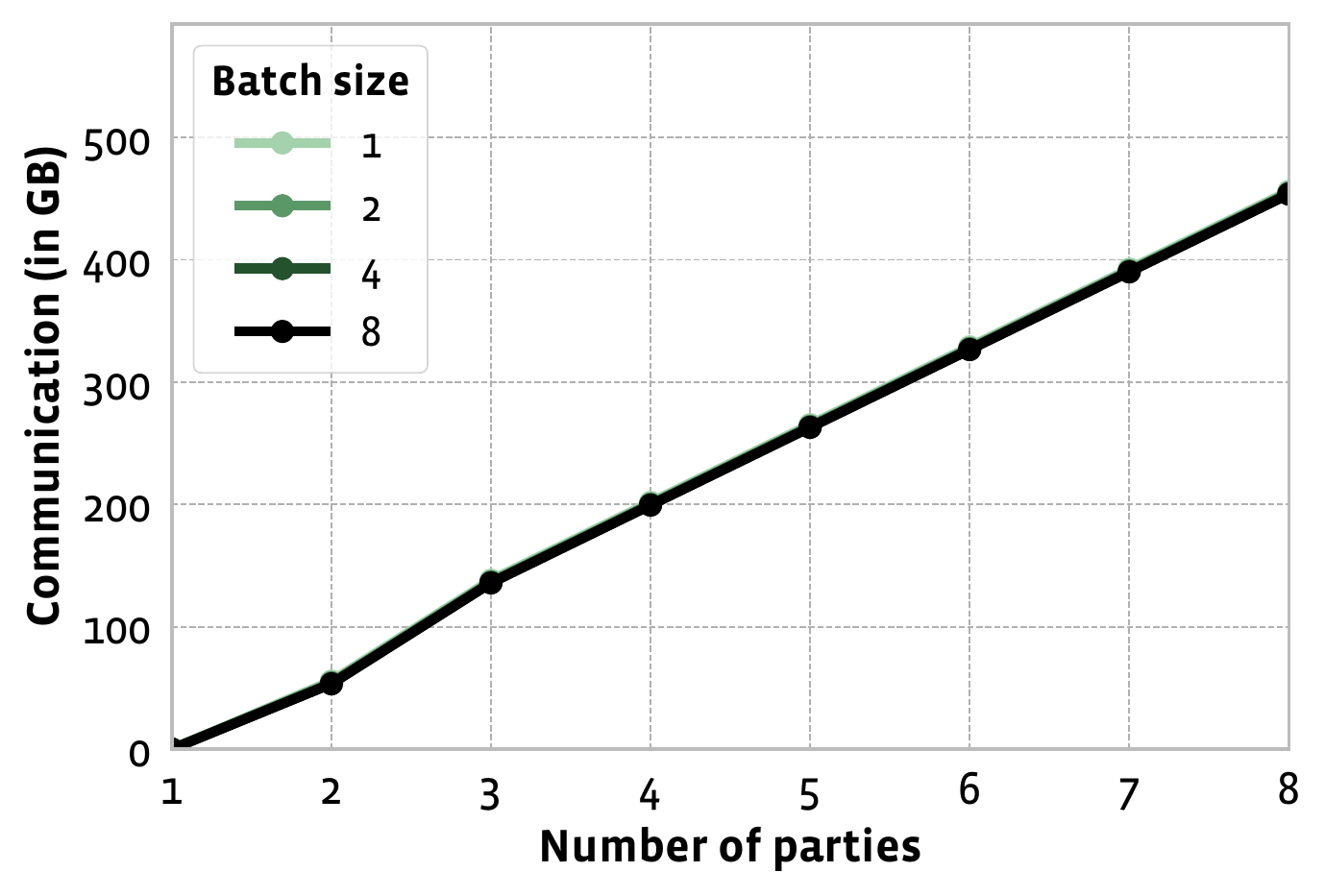}
\includegraphics[width=0.32\linewidth]{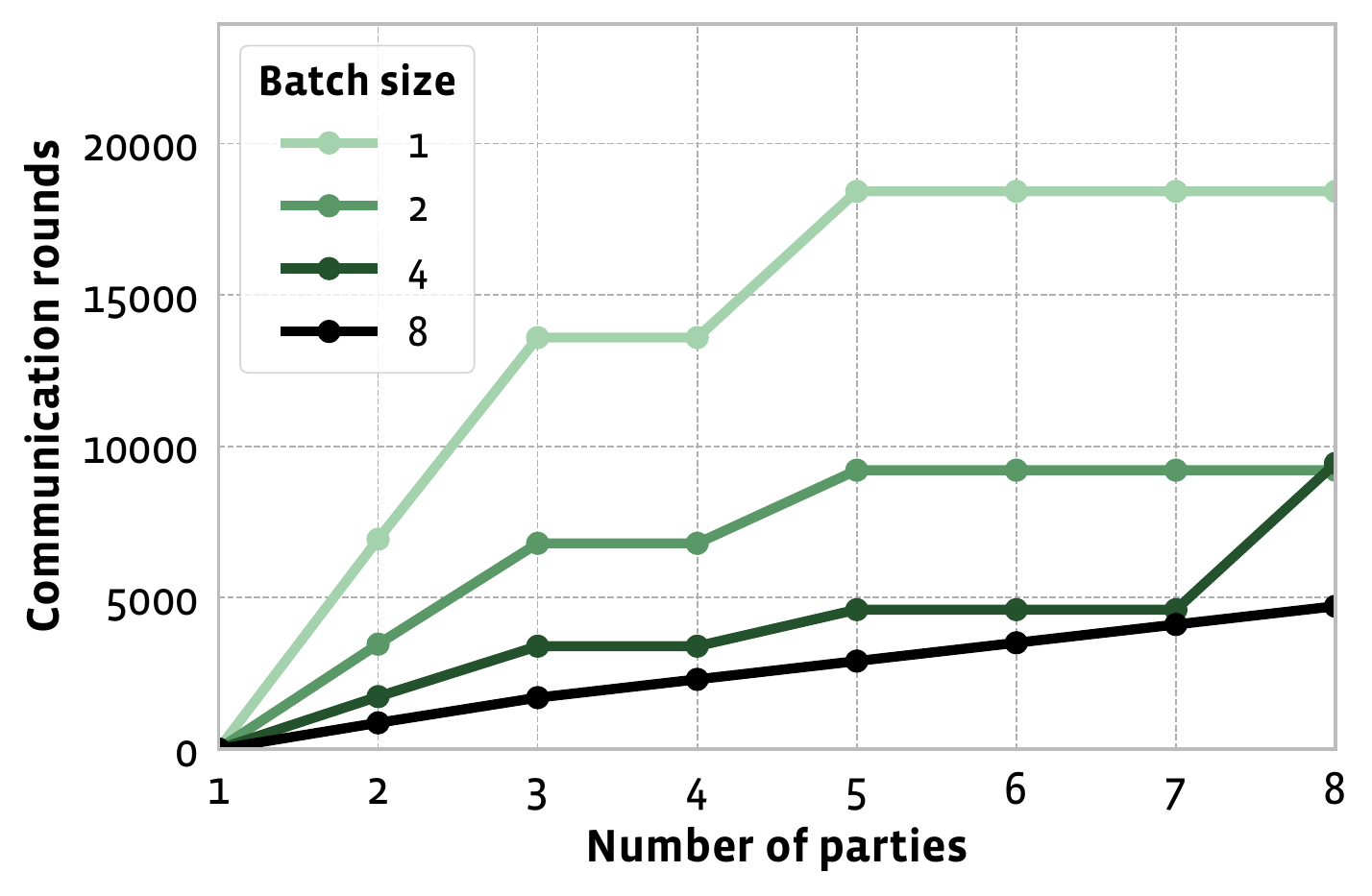}
\includegraphics[width=0.32\linewidth]{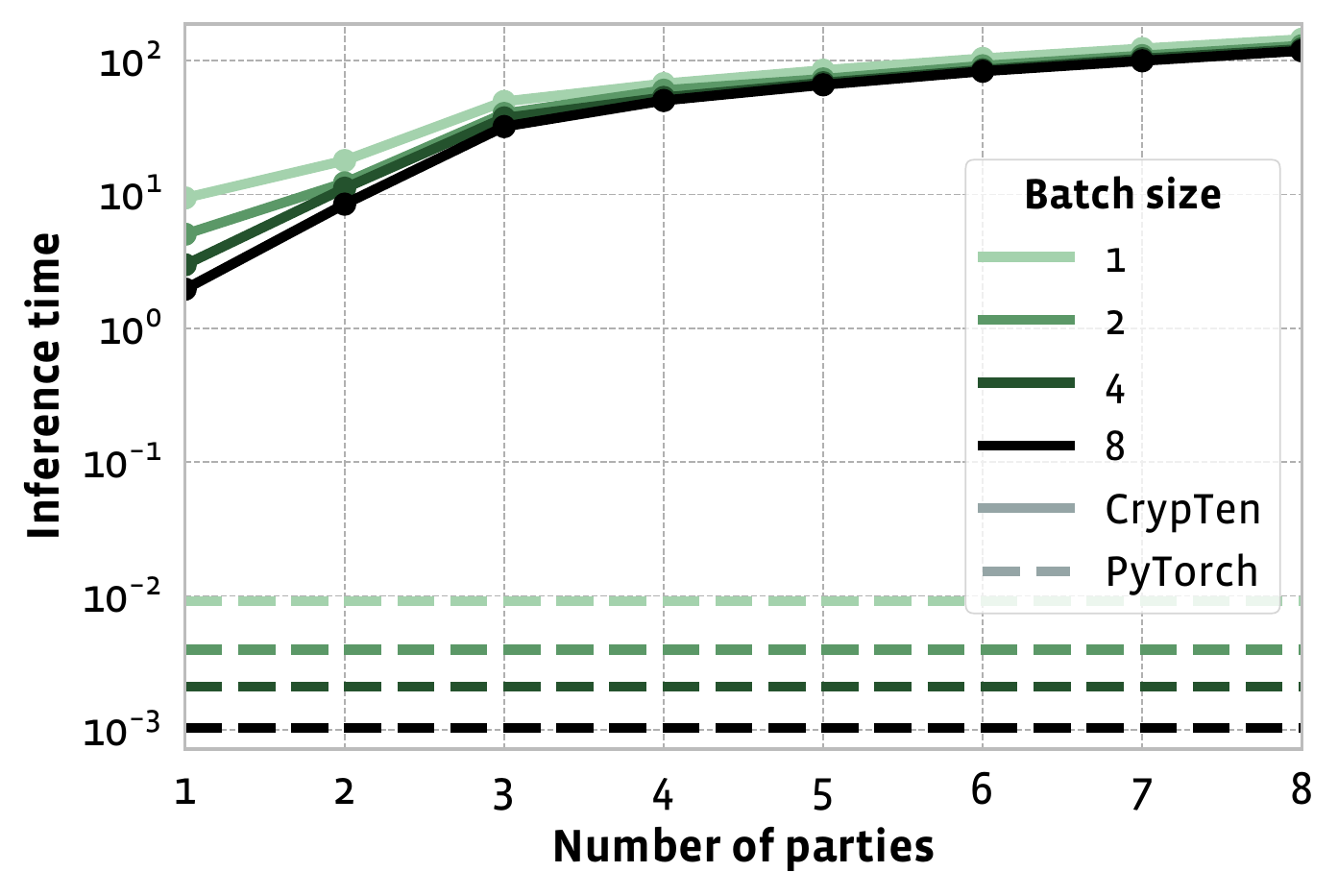}
\includegraphics[width=0.32\linewidth]{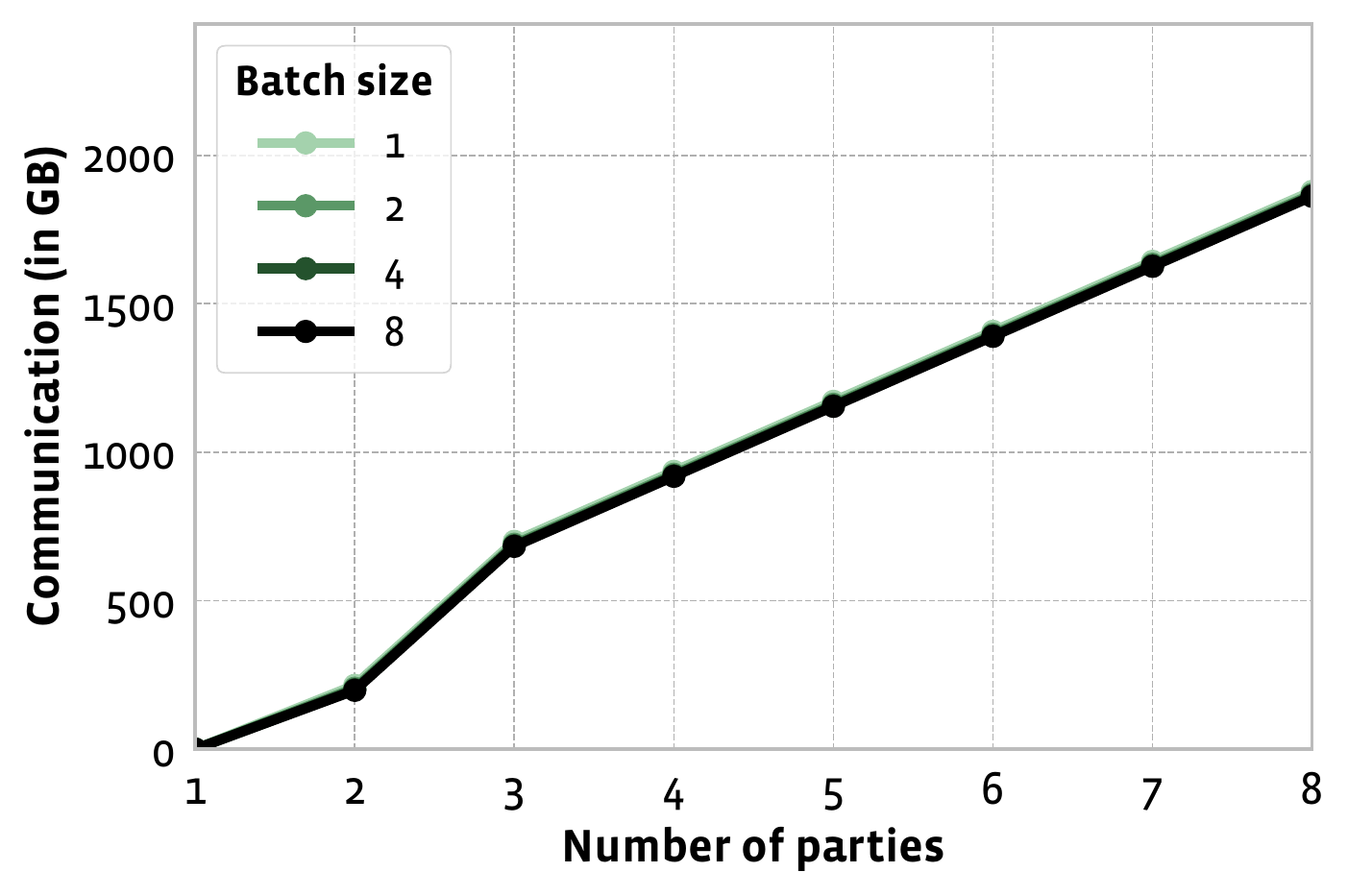}
\includegraphics[width=0.32\linewidth]{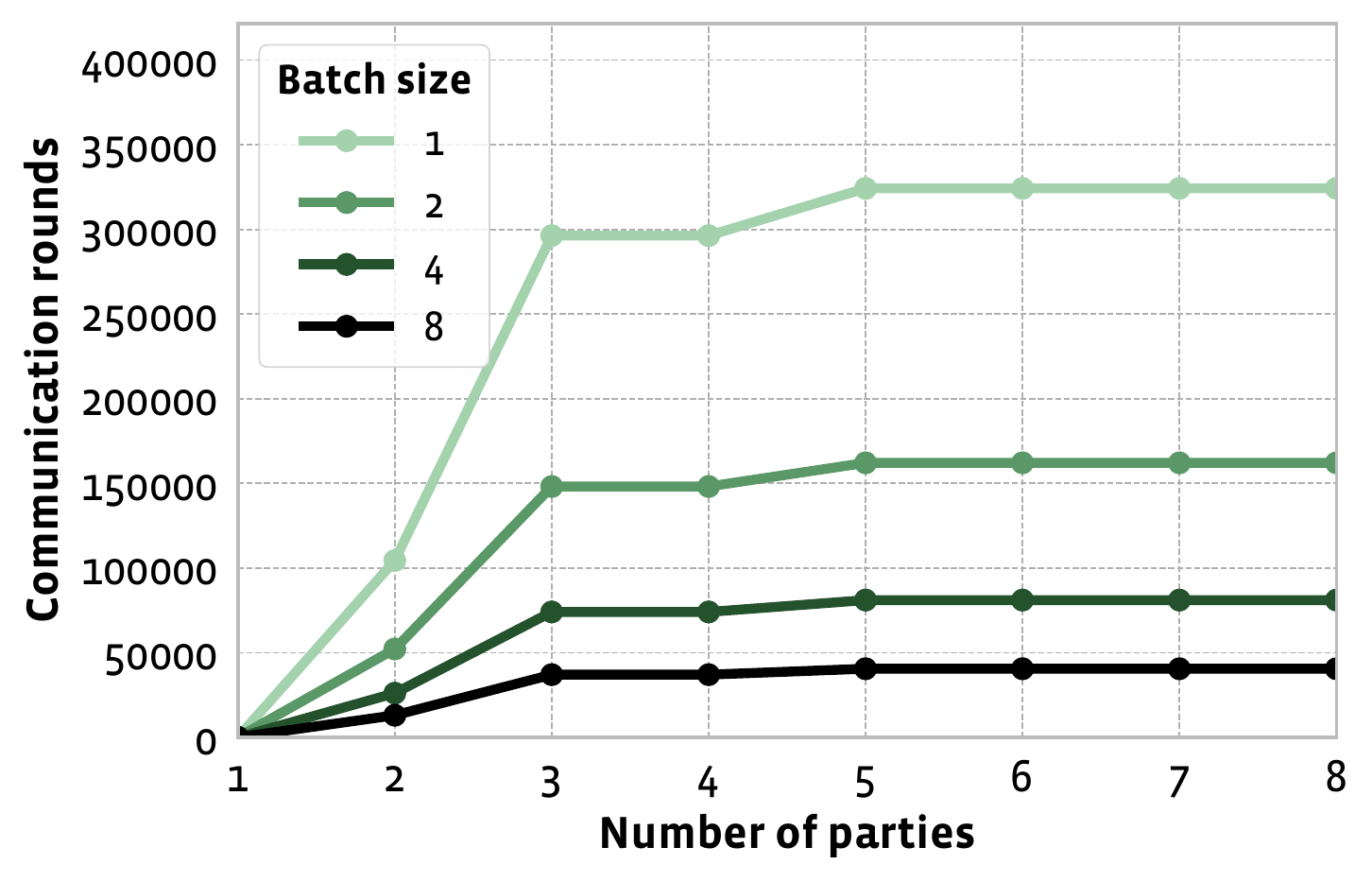}
\caption{Benchmarks for inference with image-classification models on GPUs in \crypten{} and PyTorch. \textbf{Top:} Results for ResNet-18 model. \textbf{Bottom:} Results for ViT-B/16 vision transformer.  \textbf{Left:} Average wall-clock time per sample (in seconds). \textbf{Middle:} Number of bytes communicated per sample, per party (in GB). \textbf{Right:} Number of communication rounds per sample.}
\label{fig:image_classification}
\end{figure}

\subsection{Image Classification}
We performed image-classification experiments on the ImageNet dataset using residual networks (ResNets;~\cite{he2016deep}) and vision transformers (ViT;~\cite{dosovitskiy2021}).\footnote{We adopted the ResNet implementation from \texttt{torchvision} and the ViT implementation from \url{https://github.com/rwightman/pytorch-image-models}. ViT's normalized mean squared error is larger than for other models because our Gaussian error function approximation converges slowly; see Section~\ref{sec:erf}.}
We experimented with a ResNet-18 with 18 convolutional layers and with a ViT-B/16 model that has 12 multi-head self-attention layers with 12 heads each, operating on image patches of $16 \times 16$ pixels.
Following common practice~\cite{he2016deep}, we preprocess images by rescaling them to size $256 \times 256$ and taking a center crop of size $224 \times 224$.

Figure~\ref{fig:image_classification} presents the results of our image-classification benchmarks, which show that two parties can securely evaluate a ResNet-18 model in $2.49$ seconds and a ViT-B/16 model in $8.47$ seconds.
A notable difference compared to the prior results is that the number of bytes communicated per sample is no longer reduced by increasing the batch size.
The reason for this is that the vast majority of communication involves tensors that have the same size as intermediate activation functions: activation tensors are much larger than weight tensors in image-classification models.
The amount of communication required to evaluate the ViT-B/16 model is particularly high due to the repeated evaluation of the softmax function in the attention layer of Transformers~\cite{vaswani2017attention}.
We also observe that in ResNet-18, the number of communication rounds grows faster than expected for larger batch sizes.
The reason for this is that the carry-lookahead adder~\cite{damgrd2005unconditionally} used in the conversion from $[x]$ to $\langle x\rangle$ is very memory-intensive.
When \crypten{} runs out of GPU memory, it replaces the adder by an implementation that requires $O(|\mathcal{P}|)$ communication rounds (compared to $(\log_2 |\mathcal{P}|)$ for the carry-lookahead adder) but that requires less memory.

\section{Conclusion and Future Work}
\label{sec:discussion}
In this paper, we have introduced and benchmarked \crypten{}.
We hope that \crypten{}'s flexible, machine-learning first API design and performance can help foster adoption of secure MPC in machine learning.
We see the following directions for future research and development of \crypten{}.

\noindent\textbf{Numerical issues} are substantially more common in \crypten{} implementations of machine-learning algorithms than in their PyTorch counterparts.
In particular, the fixed-point representation with $L$ bits of precision ($L\!=\!16$ by default) is more prone to numerical overflow or underflow than floating-point representations.
Moreover, arithmetic secret shares are prone to \emph{wrap-around} errors in which the sum of the shares $[x]_p$ exceeds the size of the ring, $Q\!=\!2^{64}$.
Wrap-around errors can be difficult to debug because they may only arise in the multi-party setting, in which no individual party can detect them.
We plan to implement tools in \crypten{} that assist users in debugging such numerical issues.

\noindent\textbf{End-to-end privacy} requires seamless integration between data-processing frameworks, such as secure SQL implementations~\cite{archer2018}, and data-modeling frameworks like \crypten{}.
In ``plaintext'' software, such frameworks are developed independently and combined via ``glue code'' or platforms that facilitate the construction of processing and modeling pipelines.
Real-world use cases of machine learning via secure MPC require the development of a platform that makes the integration of private data processing and modeling seamless, both from an implementation and a security point-of-view.

\noindent\textbf{Differential privacy} mechanisms may be required in real-world applications of \crypten{} in order to provide rigorous guarantees on the information leakage that inevitably occurs when the results of a private computation are publicly revealed~\cite{dwork2014algorithmic}.
\crypten{} implements sampling algorithms for the Bernoulli, Laplace, and Gaussian distributions (see appendix), which allows for the implementation of randomized response~\cite{warner1965}, the Laplace mechanism~\cite{dwork2006}, and the Gaussian mechanism~\cite{balle2018improving,dwork2014algorithmic} (although care must be taken when implementing these mechanisms~\cite{canonne2020discrete,mironov2012significance}).
In future work, we aim to use these mechanisms, for example, to do a secure MPC implementation of DP-SGD~\cite{abadi2016}.

\noindent\textbf{Threat models} may vary per use case.
Specifically, some use cases may require malicious security or may not provide a TTP.
Possible extensions may include support for malicious security via message authentication codes~\cite{spdz}, as well as support for Beaver triple generation via additive homomorphic encryption~\cite{paillier1999}, oblivious transfer~\cite{keller2016}, or more recent methods~\cite{boyle2020} to eliminate the need for a TTP.

\noindent\textbf{Model architecture design} for secure MPC is another important direction for future research.
Following prior work in this research area, this study has focused on implementing \emph{existing} machine-learning models in a secure MPC framework.
However, these models were designed based on computational considerations in ``plaintext'' implementations of the models on modern GPU or TPU hardware.
The results of our benchmarks suggest that this may be suboptimal because those considerations are very different in a secure MPC environment.
For example, the evaluation of softmax functions over large numbers of values requires a lot of communication in secure MPC, which makes attention layers very slow.
This implies that multilayer perceptron models~\cite{tolstikhin2021} are likely much more efficient than vision transformers~\cite{dosovitskiy2021,vaswani2017attention} for image classification.
We hope that \crypten{}'s machine-learning API and ease of use will spur studies that design model architectures specifically optimized for a secure MPC environment, for example, via neural architecture search~\cite{liu2018darts,mishra2020delphi,zoph2016}. 

\section{Broader Impact}
\label{sec:impact}
Although we believe that the adoption of secure MPC in machine learning can lead to the development of AI systems that are substantially more private and secure, we note that there are also potential downsides to such adoption.
In particular, because the computations in secure MPC are performed on encrypted data, it can be harder to do quality control of AI systems implemented in \crypten{}.
For example, it is impossible to inspect the values of intermediate activations (or even model outputs) unless all parties agree to reveal those values.
This may make it harder to explain why a model makes a certain decision~\cite{kim2017explainability} or to detect data-poisoning attacks~\cite{biggio2012}.
Indeed, there exist fundamental trade-offs between privacy and utility~\cite{reed1973} and those trade-offs apply to \crypten{} users, too.

It is also worth noting that, although the protocols implemented in \crypten{} come with rigorous cryptographic guarantees, practical implementations of these protocols may be broken by other means.
For example, we have no reason to assume that \crypten{} would not be susceptible to side-channel attacks~\cite{seker2020}. 
Hence, good data stewardship remains essential even when using secure computation.

\noindent\textbf{Acknowledgments.} We thank Joe Spisak, Sijun Tan, Gregory Chanan, Igor Fedan, and the PyTorch team for their support. We thank Mark Tygert, Anderson Nascimento, Amrita Roy Chowdhury, and anonymous reviewers for helpful discussions and feedback on early versions of this paper.

\bibliographystyle{abbrvnat}
\bibliography{references}


\newpage
\appendix
\begin{table*}[t]
 \begin{center}
 \rowcolors{2}{gray!25}{white}
 \begin{tabular}{l c c} 
 \toprule
 \bf MPC Primitive & \bf Round Complexity & \bf Security Proof \\
 \midrule
 \emph{Arithmetic secret sharing}   &  &  \\
 \midrule
   \hspace{5pt} Addition & 0 & Non-interactive  \\ 
   \hspace{5pt} Multiplication & 1 & \cite[Theorem 1]{sharemind2008} \\
   \hspace{5pt} Truncation & 1$^\dagger$ & Appendix~\ref{sec:arithmetic} \\
 \midrule
 \emph{Binary secret sharing} & & \\
 \midrule
   \hspace{5pt} XOR & 0 & Non-interactive \\
   \hspace{5pt} AND & 1 & \cite[\S III.B]{demmler2015aby} \\
   \hspace{5pt} Bit-shift & 0 & Non-interactive \\
 \midrule
 \emph{Conversions} & & \\
 \midrule
   \hspace{5pt} A2B & $\log_2(|\mathcal P|) \log_2(L)$ & \cite[\S 3]{damgrd2005unconditionally} \\ 
   \hspace{5pt} B2A & 1 & Appendix~\ref{sec:binary} \\
 \midrule
 \rowcolor{white}\emph{Sampling} & & \\
 \midrule
   \rowcolor{gray!25}\hspace{5pt} Bernoulli(.5) & 1 & Appendix~\ref{sec:uniform} \\
 \bottomrule
\end{tabular}
 \caption{Overview of the MPC primitives used in \crypten{}, with their round complexity and references to the relevant security proof. Round complexity is defined as the number of sequential round-trips of communication required between parties to implement a given function, using an $L$-bit ring and $|\mathcal P|$ parties. $^\dagger$The number of rounds needed for truncation in the two-party setting is zero.}
  \label{tab:security}
 \end{center}
\end{table*}

\section{Overview of Functions Implemented in \crypten{}}
\label{sec:function_overview}

Table~\ref{tab:functions_detail} gives an overview of all functions currently implemented in \crypten{}, together with a short description of the approach used to implement the function.
Random samplers are not listed in the table.
For full details on the \crypten{} secure-computation protocol, we refer to Appendix~\ref{sec:protocols}.

\begin{table}[h]
\setlength{\tabcolsep}{4pt}
\rowcolors{2}{gray!25}{white}
\resizebox{\linewidth}{!}{
\begin{tabular}{lcl}
\toprule
\bf Function & \bf Function name(s) & \bf Description\\
\midrule
Absolute & \texttt{abs} & Multiply value by its sign.\\
Addition & \texttt{add}, \texttt{+} & Each party adds their shares.\\
Argument of maximum & \texttt{argmax} & Perform pairwise comparisons or tree reduction.\\
Argument of minimum & \texttt{argmin} & Perform pairwise comparisons or tree reduction.\\
Average pooling & \texttt{avg\_pool2d} & Each party computes the average pooling of its share.\\
Batch normalization & \texttt{batchnorm} & Batch normalize values using summation, division, and variance functions.\\
Binary AND & \texttt{and}, \texttt{\&} & Compute using binary Beaver protocol.\\
Binary cross-entropy & \texttt{binary\_cross\_entropy} & Compute using logarithm, multiplication, and addition functions.\\
Binary XOR & \texttt{xor}, \texttt{\^} & Each party XORs it shares.\\
Clone & \texttt{clone} & Each party clones their share.\\
Comparison & \texttt{>=}, \texttt{<=}, \texttt{=}, \texttt{ge}, \texttt{le}, \texttt{eq} & To compare to $0$, convert to binary secret share and inspect most significant bit.\\
Concatenation & \texttt{cat} & Each party concatenates their shares.\\
\multirow{-2}{*}{Convolution} & \multirow{-2}{*}{\texttt{conv1d}, \texttt{conv2d}} & \shortstack[l]{If filter is public, each party convolves its share.\\If filter is private, compute using Beaver protocol.}\\
Cosine & \texttt{cos} & Approximate using repeated-squaring method.\\
Cross-entropy & \texttt{cross\_entropy} & Compute using softmax, logarithm, multiplication, and division functions.\\
Cumulative sum & \texttt{cumsum} & Each party computes cumulative sum of values in its share.\\
Division & \texttt{div}, \texttt{/} & If divisor is public, divide shares by value and correct for wrap-around errors.\\
Dot product & \texttt{dot} & Multiply all elements and sum results.\\
\multirow{-2}{*}{Dropout} & \multirow{-2}{*}{\texttt{dropout}} & \shortstack[l]{Each party multiplies their share with dropout mask.\\Dropout mask is not encrypted.}\\
Error function & \texttt{erf} & Approximate using Maclaurin series.\\
Exponent & \texttt{exp} & Approximate using limit approximation.\\
Flatten & \texttt{flatten} & Each party flattens their share.\\
Flip & \texttt{flip} & Each party flips their share.\\
Hard tangent & \texttt{hardtanh} & Compute using comparison, multiplication, and addition functions.\\
Logarithm & \texttt{log} & Approximate using higher-order modified Householder method.\\
Log-softmax & \texttt{log\_softmax} & Compute using exponentiation, maximum, summation, and addition functions.\\
\multirow{-2}{*}{Matrix multiplication} & \multirow{-2}{*}{\texttt{matmul}} & \shortstack[l]{If one matrix is public, each party matrix-multiplies its share.\\If both matrices are private, compute using Beaver protocol.}\\
Maximum & \texttt{max} & Compute argmax as one-hot vector; compute dot product with input.\\
Max pooling & \texttt{max\_pool2d} & Compute maximum value.\\
Mean & \texttt{mean} & Each party computes mean of its share.\\
Minimum & \texttt{min} & Compute argmin as one-hot vector; compute dot product with input.\\
\multirow{-2}{*}{Multiplication} & \multirow{-2}{*}{\texttt{mul}, \texttt{*}} & \shortstack[l]{If multiplier is public, each party multiplies its share with the\\multiplier. If multiplier is private, use Beaver protocol.}\\
Multiplexing & \texttt{where} & Multiply first value by binary mask; add second value multiplied by inverse mask.\\
Negation & \texttt{neg} & Each party negates their share.\\
Norm & \texttt{norm} & Compute using the square, sum, and square root functions.\\
Outer product & \texttt{ger} & Perform multiplication of each pair of elements.\\
Padding & \texttt{pad} &Each party pads their share.\\
Permute & \texttt{permute} & Each party permutes their share. Indexes are not encrypted.\\
Product & \texttt{prod} & Multiply all elements in the input.\\
\multirow{-2}{*}{Power} & \multirow{-2}{*}{\texttt{pow}, \texttt{pos\_pow}} & \shortstack[l]{For positive powers, multiply in log-domain and exponentiate.\\For negative powers, compute reciprocal and evaluate positive power.}\\
Reciprocal & \texttt{reciprocal} & Approximate using Newton-Rhapson iterations.\\
ReLU & \texttt{relu}, \texttt{relu6} & Compare values with $0$, and multiply values by the resulting mask.\\
Reshaping & \texttt{reshape}, \texttt{view} & Each party reshapes their share.\\
Rolling & \texttt{roll} & Each party rolls their share.\\
\multirow{-2}{*}{Scattering} & \multirow{-2}{*}{\texttt{scatter}} & \shortstack[l]{Each party scatters one share into the other share.\\Indexes are not encrypted.}\\
\multirow{-3}{*}{Selection} & \shortstack{\texttt{gather},\\ \texttt{index\_select},\\ \texttt{narrow}, \texttt{take}} & \multirow{-3}{*}{Each party selects part of their share. Indexes are not encrypted.}\\
Sigmoid & \texttt{sigmoid} & Compute using the exponential and reciprocal functions.\\
Sign & \texttt{sign} & Compare value with $0$, multiply by $2$, and subtract $1$.\\
Sine & \texttt{sin} & Approximate using repeated-squaring method.\\
Softmax & \texttt{softmax} & Compute using exponentiation, maximum, summation, and reciprocal functions.\\
Square & \texttt{square} & Compute using Beaver protocol.\\
Square root & \texttt{sqrt} & Approximate using Newton-Rhapson iterations.\\
Squeezing & \texttt{squeeze} & Each party removes dimensions with size $1$ from their share.\\
Stacking & \texttt{stack} & Each party stacks their shares.\\
Subtraction & \texttt{sub}, \texttt{-} & Each party subtracts their shares.\\
Summation & \texttt{sum} & Each party sums all values in its share.\\
Tangent & \texttt{tanh} & Perform linear transformation of sigmoid value of output.\\
Trace & \texttt{trace} & Each party sums all diagonal elements of their share.\\
Transpose & \texttt{t}, \texttt{transpose} & Each party transposes their share.\\
Unsqueezing & \texttt{unsqueeze} & Each party adds dimensions with size $1$ to their share.\\
Variance & \texttt{var} & Compute using square, addition, and subtraction functions.\\
\bottomrule
\end{tabular} 
}
\caption{Overview of all functions on tensors implemented in \crypten{}.}
\label{tab:functions_detail}
\end{table}

\section{Security of \crypten{} Functions}
\label{sec:security}
\crypten{} provides MPC implementations of a large number of functions. 
However, these functions are all composed from a small set of primitives, which are listed in Table~\ref{tab:security}.
\crypten{} provides the security guarantee in Lemma 1.  
The proof for this security guarantee follows trivially from the following observations and results from prior work:

\renewcommand{\labelenumi}{\roman{enumi}.}
 \begin{enumerate}[leftmargin=*,nosep]
   \item Operations in arithmetic secret sharing are performed in the ring $\mathbb Z_{2^L}$. Multiplications in this ring are proven to be secure in \cite{sharemind2008}.
   \item Operations in binary secret sharing are performed using the GMW protocol~\cite{gmw}. AND operations in this protocol are proven to be secure in \cite{demmler2015aby}.
   \item Conversion from arithmetic to binary (A2B) secret shares is performed using the protocol that is proven to be secure in \cite{damgrd2005unconditionally}.
   \item Tensor indexing operations like concatenation, selection, reshaping, \emph{etc.} are non-interactive, which implies an adversary cannot gain any information.
   \item Security proofs for custom MPC protocols are provided in Appendix~\ref{sec:protocols} (see Table~\ref{tab:security} for details).
   \item All other operations are compositions of secure functions (see Appendix~\ref{sec:protocols} for details). This implies they are secure because security is closed under composition~\cite{canetti2001}.
 \end{enumerate}
\newcommand\norm[1]{\left\lVert#1\right\rVert}
\section{Detailed Description of Secure MPC Protocols}
\label{sec:protocols}

\subsection{Secret Sharing}
\label{sec:sharing}
\crypten{} uses two different types of secret sharing: (1) arithmetic secret
sharing~\cite{spdz} and (2) binary secret sharing~\cite{gmw}. Below, we
describe the secret sharing methods for single values $x$ but they can
trivially be extended to real-valued vectors $\bx$.

\subsubsection{Arithmetic Secret Sharing}
\label{sec:arithmetic}
\crypten{} uses arithmetic secret sharing to perform most MPC computations. In arithmetic secret sharing, 
a scalar value $x \in \thefield$ (where $\thefield$ denotes a ring with $Q$ elements) 
is shared across $|\mathcal{P}|$ parties in such a way that the sum of the
shares reconstructs the original value $x$. We denote the sharing
of $x$ by $[x] = \{[x]_p\}_{p \in \mathcal P}$, where $[x]_p \in
\thefield$ indicates party $p$'s share of $x$. The representation has the
property that $\sum_{p \in \mathcal{P}} [x]_p \mod Q \!=\! x$. We use a fixed-point encoding to obtain 
$x$ from a floating-point value $x_R$. To do so, we multiply $x_R$ with a large scaling factor $B$ and 
round to the nearest integer: $x = \lfloor B x_R \rceil$, where $B = 2^L$ for some
precision parameter, $L$. To decode a value, $x$, we compute $x_R \approx x / B$. 
Encoding real-valued numbers this way incurs a precision loss that is inversely proportional to $L$.
Since we scale by a factor $B$ to encode numbers, we
must scale down by a factor $B$ after every multiplication. We do this using
the truncation protocol described below.

\noindent\textbf{Addition.} 
The addition of two secret-shared values, $[z] = [x] + [y]$, can be trivially implemented by having each party $p$ sum their shares of $[x]$ and $[y]$: each party $p \in \mathcal{P}$ computes $[z]_p \leftarrow [x]_p + [y]_p$. 

\noindent\textbf{Multiplication.} To facilitate multiplication of two secret
shared values, the parties use random Beaver triples \cite{beaver1991}, generated in an offline preprocessing phase.
A Beaver triple of secret shared values $([a], [b], [c])$ satisfies the property $c \!=\!
ab$. The parties use the Beaver triple to compute $[\epsilon] = [x] - [a]$ and
$[\delta] = [y] - [b]$ and decrypt $\epsilon$ and $\delta$. This does not leak
information if $a$ and $b$ were drawn uniformly at random from the ring $\thefield$.
The product $[x][y]$ can now be evaluated by computing $[c] + \epsilon[b] +
[a]\delta + \epsilon\delta$, where $\epsilon$ and $\delta$ requires a round of communication among all parties.
It is straightforward to confirm that the result of the private multiplication is correct:
\begin{align*}
\label{eq:beaver}
[c] + \epsilon[b] + [a]\delta + \epsilon\delta &= [a][b] + [x][b] - [a][b] + [y][a]-[b][a]+([x]-[a])([y]-[b])\\
    &= [x][y].\nonumber
\end{align*}
Because this result holds for any linear function, $f(\cdot)$, of two variables for which the triple $(a, b, c)$ satisfies $c = f(a, b)$, 
we use the same procedure to perform matrix multiplication and convolution.

\noindent\textbf{Square.} To compute the square $[x^2]$, the parties use a Beaver pair $([a], [b])$ such
that $b \!=\! a^2$. The parties compute $[\epsilon] = [x] - [a]$, decrypt $\epsilon$, and obtain the result via $[x^2] = [b]
+ 2\epsilon[a] + \epsilon^2$.

\noindent\textbf{Truncation.}
A simple method to divide an arithmetically shared value, $[x]$, by a public value, $\ell$, would divide the share of each party by $\ell$. However, such a method can produce incorrect results when the sum of shares ``wraps around'' the ring size, $Q$. Defining $\theta_x$ to be the number of wraps such that $x = \sum_{p \in \mathcal P} [x]_p - \theta_x Q$, indeed, we observe that:
\[
    \frac{x}{\ell} = \sum_{p \in \mathcal P} \frac{[x]_p}{\ell} - \frac{\theta_x}{\ell} Q \ne \sum_{p \in \mathcal P} \frac{[x]_p}{\ell} - \theta_x Q.
\]
Therefore, the simple division method fails when $\theta_x \ne 0$, which happens with probability $\nicefrac{x}{Q}$ in the two-party case. Many MPC implementations
specialize to the $|\mathcal P| =2$-party case and assume this probability is
negligible~\cite{mohassel2017secureml, sadegh2017chameleon, wagh2018securenn}.
However, when $|\mathcal P|>2$ the probability of failure grows rapidly and we must account
for the number of wraps, $\theta_x$.
\begin{algorithm}[tb]
   \caption{Private computation of the wrap count for an arithmetically shared value.}
   \label{alg:wraps}
\begin{algorithmic}
   \STATE {\bfseries Input: } 
   {Arithmetic secret shared value $[x]$,\\
   Secret shared random value $[r]$ and its wrap count $[\theta_x]$.} \\
   \vspace{2mm}
   \STATE{Compute: $[z] \leftarrow [x] + [r]$}
   \FOR{$p \in \mathcal P$}
   \STATE Party $p$ computes: $[\beta_{xr}]_p \leftarrow ([x]_p + [r]_p - [z]_p) / Q$.
   \ENDFOR
   \STATE Construct: $[\beta_{xr}] = \{[\beta_{xr}]_p\}_{p \in \mathcal P}$
   \STATE Decrypt: $z \leftarrow$ reveal($[z]$)
   \STATE Compute during decryption: $\theta_z \leftarrow  (\sum_p [z]_p - z) / Q$.
   \STATE Compute: $[\eta_{xr}] \leftarrow z < [r]$
   \STATE Compute: $[\theta_x] \leftarrow \theta_z + [\beta_{xr}] - [\theta_r] - [\eta_{xr}]$
\end{algorithmic}
\end{algorithm}
We do so by privately computing a secret share of the number of wraps in $x$, $[\theta_x]$. To this end, we
define three auxiliary variables:
\begin{itemize}[leftmargin=*]

    \item $\theta_x$ represents the number of wraps produced by the shares of a secret shared variable $[x]$, such that
        $x = \sum_p [x]_p - \theta_x Q$, where $Q$ is the ring size.

    \item $\beta_{xr}$ represents the differential wraps produced between each party's shares of two secret shared variables,
        $[x]$ and $[r]$, such that $[x]_i + [r]_i \mod Q = [x]_i + [r]_i - [\beta_{xr}]_i Q$. 

    \item $\eta_{xr}$ represents the wraps produced by two plaintext variables, $x$ and $r$, such that 
        $x + r \mod Q = x + r - \eta_{xr} Q$.
\end{itemize}

We use these variable in Algorithm~\ref{alg:wraps} to compute $[\theta_x]$. This approach is inspired by Algorithm 4 of \cite{wagh2018securenn}, but extends to an arbitrary number of parties.
The correctness of this algorithm can be shown through the following reduction:
\begin{equation*}
    \begin{array}{rcl}
        z & = & x + r - \eta_{xr} Q\\
        \sum_p [z]_p - \theta_z Q & = & (\sum_p [x]_p - \theta_x Q) + (\sum_p [r]_p - \theta_r Q) - \eta_{xr} Q\\
        \sum_p [z]_p - \theta_z Q & = & (\sum_p [x]_p + [r]_p) - (\theta_x + \theta_r + \eta_{xr}) Q\\
        \sum_p [z]_p - \theta_z Q & = & (\sum_p [z]_p - [\beta_{xr}]_p Q) - (\theta_x + \theta_r + \eta_{xr}) Q\\
        \sum_p [z]_p - \theta_z Q & = & (\sum_p [z]_p) - (\beta_{xr} + \theta_x + \theta_r + \eta_{xr}) Q\\
        \theta_x & = & \theta_z + \beta_{xr} - \theta_r - \eta_{xr}.
    \end{array}
\end{equation*}

We then use $[\theta_x]$ to correct the value of the division by $\ell$:
\[
    \frac{x}{\ell}  = [y] - [\theta_x] \frac{Q}{\ell} \quad \text{where} \quad
    [y] = \left\{\frac{[x]_p}{\ell}\right\}_{p\in\mathcal P}.
\]

In practice, it can be difficult to compute $[\eta_{xr}]$ in Algorithm~\ref{alg:wraps}.
However, we note that $\eta_{xr}$ has a fixed probability of being non-zero, irrespective of the number of parties.
Indeed, regardless of the number of parties, we have $P(\eta_{xr} \neq 0) = \nicefrac{x}{Q}$. In practice, we can therefore skip the
computation of $[\eta_{xr}]$ and simply set $\eta_{xr}=0$. This implies that incorrect results can be produced
by our algorithm with small probability. For example, when we encode a real value $\hat
x$ using a fixed-point encoding $x = \lfloor B \hat x \rceil$, truncation will produce  an error with
probability $P(\eta_{xr} \neq 0) = \nicefrac{\lfloor B \hat x \rceil}{Q}$. This
probability can be reduced by increasing $Q$ or reducing the precision
parameter, $B$. 

\underline{\emph{Security proof.}} One can show the security of Algorithm~\ref{alg:wraps} by noting that the only information gained by an adversary is the revealed shares of $[z]$, which are indistinguishable from white uniform random noise because shares of $[r]$ are chosen to be uniformly random.

\subsubsection{Binary Secret Sharing}
\label{sec:binary}
Binary secret sharing is a special case of arithmetic secret sharing that operates within the binary field $\mathbb{Z}/2\mathbb{Z}$.
In binary secret sharing, a sharing $\langle x \rangle$ of a value $x$ is generated as a set of arithmetic secret shares of
the bits of $x$ within the binary field. Each party $p \in \mathcal P$  holds a share $\langle x\rangle_p$ that satisfies 
$x = \bigoplus_{p \in \mathcal P} \langle x \rangle_p $. Because addition and multiplication modulo 2 are equivalent to binary 
XOR and AND operations, we can use bitwise operations on integer types to vectorize these operations. 

Note that XOR and AND operations form a basis for the set of Turing-complete operations (via circuits). However each
sequential AND gate requires a round of communication, which makes all but very simple circuits very inefficient to evaluate via binary secret sharing.
In \crypten{}, we only use binary secret sharing to implement comparators.

\noindent\textbf{Bitwise XOR.} Similar to addition in arithmetic secret sharing, a binary XOR of two binary secret-shared values, 
$\langle z \rangle =\langle x \rangle + \langle y \rangle$ can be trivially implemented by having each party XOR their shares of $\langle x\rangle$ and $\langle y\rangle$. 
That is, each party $p \in \mathcal{P}$ computes $\langle z \rangle_p \leftarrow \langle x\rangle_p \oplus \langle y\rangle_p$. 

\noindent\textbf{Bitwise AND.} Since the bitwise AND operation is equivalent multiplication mod 2, we can utilize the same method we use to multiply arithmetic secret shared values. 
To facilitate bitwise AND of two binary secret-shared values, the parties use random triples generated in an offline preprocessing phase.
The generated triple $(\langle a \rangle, \langle b \rangle, \langle c\rangle)$ satisfies the property $c \!=\!a\otimes b$. The parties then compute 
$\langle\epsilon\rangle = \langle x\rangle \oplus \langle a\rangle$ and
$\langle\delta\rangle = \langle y\rangle \oplus \langle b\rangle$ and decrypt $\epsilon$ and $\delta$. This does not leak
information since $a$ and $b$ contain bits drawn uniformly at random.
 $\langle x\rangle \otimes \langle y\rangle$ can now be evaluated by computing $\langle c \rangle \oplus (\epsilon \otimes\langle b\rangle) \oplus
(\langle a\rangle\otimes\delta) \oplus (\epsilon\otimes\delta)$.
Correctness follows from the same logic as multiplication in arithmetic secret sharing.
We note that revealing $\epsilon$ and $\delta$ requires a round of communication among all parties in this protocol.

\noindent\textbf{Logical shifts.} Because each bit of a binary secret-shared value is an independent secret-shared bit, logical shifts can be performed trivially.
To shift the bits of a binary secret-shared value $\langle x \rangle$ by a constant $k$, each party can compute the shift locally on its share, $\langle y \rangle_p = \langle x \rangle_p >\!> k$.

\subsubsection{Converting Between Secret-Sharing Types}
Many machine-learning models require both functions that are easier to compute on arithmetic secret shares (\emph{e.g.}, matrix multiplication) and functions that are easier to implement via circuits on binary secret shares (\emph{e.g.}, argmax). Therefore, \crypten{} uses both types of secret sharing and converts between the two types as needed using the techniques proposed in~\cite{demmler2015aby}.

\noindent\textbf{From $[x]$ to $\langle x\rangle$:} To convert from an arithmetic
share $[x]$ to a binary share $\langle x \rangle$, each party first secretly
shares its arithmetic share with the other parties and then performs addition
of the resulting shares. The parties construct binary secret shared values $\langle y_p \rangle$ where each $y_p$ represents one of the arithmetic secret shares $y_p = [x]_p$.  
This process is repeated for each party $p \in \mathcal{P}$ to create
binary secret shares of all $|\mathcal{P}|$ arithmetic shares $[x]_p$. Subsequently, the
parties compute $\langle x \rangle = \sum_{p\in \mathcal P} \langle y_p\rangle$. To compute the sum, a carry-lookahead adder circuit can be evaluated in $\log_2 (|\mathcal P|)\log_2(Q)$
rounds~\cite{catrina2010improved, damgrd2005unconditionally}.
In practice, the carry-lookahead adder circuit is quite memory-intensive.
When \crypten{} runs out of GPU memory, we adopt an alternative adder circuit that requires substantially less memory but performs $|\mathcal P| \log_2(L)$ communication rounds to perform the summation.

\noindent\textbf{From $\langle x\rangle$ to $[x]$:} To convert from a binary share
$\langle x \rangle$ to an arithmetic share $[x]$, the parties compute $[x] =
\sum_{b=1}^B 2^b \left[\langle x\rangle^{(b)}\right]$, where $\langle x\rangle^{(b)}$
denotes the $b$-th bit of the binary share $\langle x\rangle$ and $B$ is the total number of bits in the shared secret. To create the arithmetic
share of a bit, the parties use $b$ pairs of secret-shared bits $([r], \langle r \rangle)$ generated offline.
Herein, $[r]$ and $\langle r \rangle$ represent
arithmetic and binary secret-shares of the same bit value $r$. Parties then use Algorithm~\ref{alg:B2A} to generate
$\left[\langle x \rangle^{(b)}\right]$ from $\langle x \rangle^{(b)}$. This process can be performed for each bit in parallel, reducing the 
number of communication rounds required for the conversion process to one. 

\underline{\emph{Security proof.}} One can show the security of Algorithm~\ref{alg:B2A} by noting that  the only information gained by an adversary is the revealed shares of $\langle z\rangle$, which are indistinguishable from white Bernoulli random noise because shares of $\langle r\rangle$ are chosen to be uniformly random.

\begin{algorithm}[tb]
   \caption{Private single bit conversion from binary to arithmetic sharing.}
   \label{alg:B2A}
\begin{algorithmic}
   \STATE {\bfseries Input: } 
   {Binary secret shared bit $\langle b\rangle$; random bit in both arithmetic and binary sharing $[r], \langle r\rangle$.} \\
   \vspace{2mm}
   \STATE{Compute: $\langle z\rangle \leftarrow \langle b\rangle \oplus \langle r\rangle$}.
   \STATE Decrypt: $z \leftarrow$ reveal($\langle z \rangle$).
   \STATE Compute: $ [b] \leftarrow  [r] + z - 2[r]z$.
\end{algorithmic}
\end{algorithm}

\subsubsection{Logic-based Operations}
\label{sec:logic}

Many applications require implementations of logic-based operators to make branching decisions and
compute piece-wise functions.

\noindent\textbf{Comparisons.} To compare two secret-shared values $[x]$ and $[y]$, we can produce $[x < y]$ by computing their difference $[z] = [x] - [y]$ and comparing the result to zero: $[z<0]$.
We compute $[z<0]$ by first converting $[z]$ to a binary secret-share $\langle z \rangle$, computing its sign bit using a right shift $\langle b \rangle = \langle z \rangle >\!> (L - 1)$, and converting
the resulting bit to an arithmetic sharing $[b]$. Because we are using an integer encoding, the most significant bit of $z$ represents its sign. It is possible to compare $[x < y]$ directly using 
a less-than circuit, but this requires converting an extra value to binary secret sharing and incurring another $\log_2 L$ rounds of communication to compute the less-than circuit.

We can use the ability to compute $[x<y]$ to compute all other comparators on $[x]$ and $[y]$:
\begin{align*}
     [x > y] &= [y < x]\\
     [x \ge y] &= 1 - [x < y]\\
     [x \le y] &= 1 - [y < x]\\
     [x = y] &= [x \le y] - [x < y]\\
     [x \ne y] &= 1 - [x = y].
\end{align*}
We optimize evaluation of the is-equal operator by computing $[x \le y]$ and $[x < y]$ in parallel.

\noindent\textbf{Multiplexing.} Multiplexing is a very valuable tool for computing conditional and piece-wise functions. To multiplex between two values $[x]$ and $[y]$ based on a condition $c$,
we must first evaluate $c$ to a a binary value $[c]\in \{[0],[1]\}$. We can then compute $[c~?~x : y] = [c][x] + (1 - [c])[y]$. This allows us to evaluate if-statements using \crypten{}, where $[x]$ is the result
when the if-statement is executed, and $[y]$ is the result otherwise. However, unlike if-statements, both results must be evaluated, meaning we cannot use tree-based or dynamic programming
techniques to optimize algorithm runtimes.

\noindent\textbf{Sign, absolute value, and ReLU.} Several important functions can be computed using the multiplexing technique. We can compute $\operatorname{sign}([x]) = 2[x > 0] - 1$. We can then use
this to compute $|[x]| = [x]\operatorname{sign}([x])$. Similarly we can compute the ReLU function by noting $\operatorname{ReLU}([x]) = [x][x > 0]$.

\noindent\textbf{Argmax and maximum.}
\crypten{} supports two methods for computing maximums $[\max x]$. Both methods first compute a one-hot argmax mask that contains a one at the index containing a maximal element $[y] = \operatorname{argmax}([x])$. A maximum can then be obtained by taking the sum $[\max x] = \sum_i [y_i][x_i]$ where the sum is taken along the dimension over which the maximum is being computed. By default, the argmax is computed using a tree-reduction algorithm, though configurations are available to use pairwise comparisons depending on network bandwidth / latency.

The \textit{tree-reduction} algorithm computes the argmax by partitioning the input into two halves, then comparing the elements of each half. This reduces the size of the input by half in each round, requiring $O(\log_2 N)$ rounds to complete the argmax. This method requires order $O(\log_2 N)$ communication rounds, $O(N^2)$ communication bits, and $O(N)$ computation complexity.

The \textit{pairwise} method generates a matrix $[A]$ whose rows are constructed by the pairwise differences of every pair of elements $\forall i \neq j: [A_{ij}] = [x_i - x_j]$. We then evaluate all comparisons simultaneously by computing $[A \ge 0]$. All maximal elements will correspond to columns whose elements are all greater than 0, so we can compute the argmax mask $[m]$ by taking the sum over all columns of $[A]$. However, if more than one maximal element exists, this will result in a mask $[m]$ that is not one-hot. To make this one-hot we take a cumulative sum $[c]$ of $[m]$ and return $[c < 2][m]$ to return the index of the first maximal element. This method requires $O(1)$ communication rounds, $O(N^2)$ communication bits, and $O(N^2)$ computation complexity. In theory, because of constant-round communication, this method should be more efficient than the tree-reduction algorithm when the network latency is high.

\noindent\textbf{Argmin and minimum.} To compute minimums and argmins, we compute our argmax mask with a negated input: $[\operatorname{argmin} x] = [\operatorname{argmax} (-x)]$.

\subsection{Mathematical Approximations}
\label{sec:approximations}

Many functions are very expensive to compute exactly using only addition, multiplication, truncation, and comparisons. \crypten{} uses numerical approximations to compute these functions, optimizing for accuracy, domain size, and efficiency when computed on secret shares. Each of these approximations has a specific domain over which the approximation converges well. One can modify the domain of convergence for certain functions using function-specific identities. For example, $\forall a\in \mathbb{R}$:
\begin{align*}
	\ln(x) &= \ln(ax) - \ln(a)\\
	x^{-1} &= a(ax)^{-1}\\
	e^x &= e^{x-a}e^{a}.
\end{align*}
\crypten{} also offers configurable parameters for protocol-specific optimizations, for example, custom initializations that improve convergence  for iterative methods in a pre-specified input domain.

\subsubsection{Exponential, Sine, and Cosine}
\label{exponential}

There are many well-known polynomial approximations for the exponential function, for example, the Taylor series, $ e^x = \sum_{n=0}^{\infty} \frac{1}{n!}x^n $. However, because exponentials grow much faster than polynomials, the degree of the polynomial we would need to approximate the exponential function increases exponentially as the domain increases. Therefore, we instead use the limit approximation, which allows us to do repeated squaring very efficiently:
$$ e^x = \lim_{n\rightarrow\infty} \left(1 + \frac{x}{2^n}\right)^{2^n}.$$


\crypten{} can also use the repeated-squaring method to compute complex exponentials efficiently, which enables the computation of the sine and cosine functions:
\begin{align*}
 \cos x &= \Re(e^{ix})\\
 \sin x &= \Im (e^{ix}).
\end{align*}

\subsubsection{Reciprocal}
\label{reciprocal}

\crypten{} uses Newton-Raphson iterations to compute the reciprocal function. This method uses an initial guess, $y_0$, for the reciprocal and repeats the following update:
$$ y_{n+1} = y_n(2 - xy_n).$$
This will converge to $\lim_{n\rightarrow\infty}y_n \!=\! \frac{1}{x}$ quadratically as long as the initial guess $y_0$ meets the Newton-Raphson convergence criterion, which is $0 \!<\! y _0 \!<\! \frac{2}{x}$ for the above. By default, \crypten{} uses: $$y_0(x) = 3e^{0.5 - x} + 0.003,$$ to initialize the approximation, which provides convergence on a large domain. This function was found by inspection and can be replaced by a user-defined value using \crypten{}'s configuration API. Because this method only converges for positive values of $x$, we compute the reciprocal using the identity $\frac{1}{x} = \frac{\operatorname{sgn}x}{|x|}$. (Note that square matrix inverses and Moore-Penrose inverses can be found using similar techniques given input matrices with singular values that meet the convergence criterion.)

\subsubsection{Square Root and Normalization}
\label{sqrt}

\crypten{} uses Newton-Raphson iterations to compute square roots. However, the Newton-Raphson update formula for square roots, $ y_{n+1} =\frac{1}{2}(y_n + \frac{x}{y_n})$ is quite inefficient to compute on secret shares. Instead, we use the much more efficient Newton-Raphson update formula for inverse square root: $$ y_{n+1} = \frac{1}{2}y_n(3 - xy_n^2).$$ We then multiply by the input $x$ to obtain the square root: $\sqrt{x} = (x^{-0.5})x$. We can also use the inverse square root function to efficiently normalize values via: $\frac{x}{\norm{x}} = x \left(\sum_i x_i^2\right)^{-\nicefrac{1}{2}}$.

\subsubsection{Logarithm and Exponents}
\label{logarithm}

To compute logarithms, \crypten{} uses higher-order iterative methods to achieve better convergence. The following update formula can be found using high-order modified Householder methods on $\ln(x)$ \cite{householder} or by manipulating the Taylor series expansion of $\ln(1-x)$:
\begin{align*}
h_n &= 1 - xe^{-y_n}\\
y_{n+1} &= y_n - \sum_{k=1}^{\infty} \frac{1}{k}h_n^k.
\end{align*}
Note that at each step $\ln x = y_n + \ln (1 - h_n)$, but we can only approximate $\ln(1-h_n)$ using a truncated Taylor Series approximation. For this method, the order of the Householder method (\emph{i.e.}, the polynomial degree in the second equation) will determine the speed of convergence. Since the convergence rate per iteration increases proportionally to the degree of the polynomial, whereas an exponential must be computed for each iteration, it is more computationally efficient to use high-degree polynomials instead of doing many iterations. By default, \crypten{} uses a polynomial of degree $8$, the initialization $y_0 = \frac{x}{120} - 20e^{-2x - 1 } + 3$, and $3$ iterations. This provides effective convergence on the domain $[10^{-4}, 10^2]$.

Using the logarithm and exponential functions, we can also compute arbitrary public or private exponents on positive inputs $x$ using the equation $x^y = e^{y\ln(x)}$. 

\subsubsection{Sigmoid and Hyperbolic Tangent}
\label{sigmoid}

We have explored several methods for computing logistic functions in MPC, including direct computation, rational approximations, and Chebyshev polynomial approximations \cite{guo2020}. We have found that direct computation is the most efficient when it is combined with some specific optimizations. Specifically, \crypten{} uses the exponential and reciprocal functions to compute:
$$\sigma(x) = \frac{1}{1 + e^{-x}}.$$
We optimize this computation by noting that the range of the sigmoid function is $[0, 1]$, and the range for the positive half of its domain is $[0.5, 1]$. Therefore, when we compute the reciprocal using the method described in Section~\ref{reciprocal}, we compute $\sigma(|x|)$ using an initialized value of $0.75$ for the Newton-Raphson iterations to improve convergence. We extend the result to the full domain by noting $\sigma(-x) = 1 - \sigma(x)$. We compute the hyperbolic tangent function via $\tanh(x) = 2\sigma(2x) - 1$.

\subsubsection{Gaussian Error Function}
\label{sec:erf}
We use a Maclaurin series to approximate the Gaussian error function $\textrm{erf}(x) = \nicefrac{2}{\sqrt{\pi}} \int_0^x e^{-x^2} dx$. 
The resulting approximation is given by:
$\textrm{erf}(x) \approx \frac{2}{\sqrt{\pi}} \sum_{k=0}^K \frac{(-1)^k x^{2k+1}}{k!(2k+1)}$,
where $K$ is the number of terms in the approximation (we set $K=8$ by default).
Although the approximation works reasonably well in practice, we note that it is known to have poor convergence when $x > 1$ (see OEIS A007680).

\subsection{Random Sampling}
\label{sec:sampling}

Several applications of privacy-preserving computations require secret-shared generation of random numbers such that
no party can gain any information about the value of realizations. We use the following methods for generating secret shares
of random samples from several popular distributions.

\subsubsection{Uniform Sampling}
\label{sec:uniform}
Due to quantization introduced by our encoding with scale $2^L$, we can only produce discrete uniform random variables
with $2^L$ possible values. To do so, we produce samples $[u]\sim Uniform(0,1)$
by generating $L$ bits as Rademacher variates. To generate these bits,
each party randomly generates its own binary secret-share with the same distribution locally. 
The XOR sum of independently distributed Rademacher variates, $u = \oplus_{p\in\mathcal P}\langle u\rangle_p$, 
is itself a Rademacher variate and is uncorrelated with any of the input bits.

\underline{\emph{Security proof.}} One can show the security of this sampler by noting that no adversary could gain any information about the sampled bit from its own binary share of the bit, because the XOR sum  of independently distributed Rademacher variates is uncorrelated with any of the input bits. 
The bits are then converted to an arithmetic share $[u]$ using Algorithm~\ref{alg:B2A}, which is itself secure.

\subsubsection{Bernoulli Sampling}
To compute a Bernoulli random variable with arbitrary mean $[b] \sim Bern(p)$, we first generate a uniform random variable
$[u] \sim Uniform(0, 1)$ and compute $[b] = [u > p]$. Note that due to quantization in $[u]$, the true probability parameter of the Bernoulli random sample
is quantized to the nearest multiple of $2^{-L}$, as would have happened if $p$ was encoded using the fixed-point encoder.

\subsubsection{Gaussian Sampling}
Gaussian random samples $[x]\sim \mathcal{N}(\mu, \sigma^2)$ can be computed using the Box-Muller transform. Given a pair of independent uniformly
distributed random variables $([u_1], [u_2])$, two independent Gaussian random variables $([x_1], [x_2])$ from $\mathcal{N}(0, 1)$ can be generated
by computing:
$$ [x_1] = \sqrt{-2\ln [u_1]} \cos (2\pi[u_2])$$
$$ [x_2] = \sqrt{-2 \ln [u_1]} \sin (2\pi[u_2]).$$

Since the range of the uniform inputs is $[0, 1]$, we optimize our numerical approximations for better performance on this domain.  To obtain samples $[y] \sim \mathcal{N}(\mu, \sigma^2)$, we compute $[y] = \sigma [x] + \mu$.

\subsubsection{Exponential and Laplace sampling}
Exponential random variables $[x] \sim Exp(\lambda)$ can be computed using the inverse CDF method. Given a uniform random sample
$[u] \sim U[0, 1]$, an exponential random variable is generated via:
$$ [x] = -\lambda^{-1} \ln([u]).$$

Again, we optimize the logarithm for the domain [0, 1]. 

A Laplace distributed random sample $[y] \sim Lap(\mu, k)$ can be generated from an exponential random sample, $[x] \sim Exp(k^{-1})$, and a
Rademacher variate, $[b]$, by evaluating $[y] = (2[b]- 1)[x]$.

\subsubsection{Weighted Random Sampling}
To produce a weighted random sample of inputs $[x_i]$ with weights given by $[w_i]$, we first generate a uniform random sample in $\left([0, \sum_i [w_i]\right)$ by drawing a uniform sample, $[u]$, and evaluating $[r] = [u]\left[\sum_i w_i \right]$. Care should be taken to avoid precision issues caused by generating $[u]$ in fixed-point with finite precision. We then compute the cumulative sum values $[c_i]$ of the weights $[w_i]$, and compare those values to our random value $[m_i] = [c_i > r]$. This produces a mask vector whose entries are all zero below some index $j$ and all one above index $j$. To convert this mask vector into a one-hot vector, we append a zero in front of the $[m_i]$-values and compute $[o_i] = [m_i] - [m_{i+1}]$. Finally, we obtain the selected sample from the inputs $[x_i]$ by multiplying the samples with the one-hot vector and summing: $[y] = \sum_i [x_i][o_i]$.



\section{Comparison with Secure MPC Frameworks for Machine Learning}
\label{sec:framework_comparison}
Table~\ref{tab:framework_comparison} presents a comparison of \crypten{} with other secure MPC frameworks for machine learning.
For each framework, the table shows whether the framework supports maliciously secure threat models, can generate Beaver triples (if needed) without requiring a trusted third party, supports GPU computations, supports model training, supports general purpose function evaluation, and implements automatic differentiation (autograd).
We define a secure MPC framework for machine learning to be general-purpose if it supports at least the following functions: linear functions, convolutions, rectified linear units (ReLU), max-pooling, and batch normalization.\footnote{\crypten{} supports a variety of functions beyond these five functions, but we focus on these five in our comparison as they are the main building blocks of many deep network architectures.}

\begin{table}[h]
\setlength{\tabcolsep}{4pt}
\rowcolors{2}{gray!25}{white}
\resizebox{\linewidth}{!}{
\begin{tabularx}{1.2\linewidth}{l| *6{>{\Centering}X}}
\toprule
\rowcolor{white} \bf \multirow{2}{*}{Framework} & \bf Malicious security & \bf Triple generation & \bf Supports GPUs & \bf Supports training & \bf General purpose$^\dagger$ & \bf Supports autograd\\
\midrule
\multicolumn{7}{l}{\textit{Two parties}}\\
\midrule
Chameleon~\cite{sadegh2017chameleon} & \redcross & \redcross & \redcross & \redcross & \redcross & \redcross\\
Delphi~\cite{mishra2020delphi} & \redcross & \ForestGreencheck & \redcross & \redcross & \redcross & \redcross\\
EzPC~\cite{chandran2019} & \redcross & \ForestGreencheck & \redcross & \redcross & \redcross & \redcross\\
Gazelle~\cite{juvekar2018gazelle} & \redcross & \ForestGreencheck & \redcross & \redcross & \redcross & \redcross\\
MiniONN~\cite{liu2017minionn} & \redcross & \ForestGreencheck & \redcross & \redcross & \redcross & \redcross\\
PySyft~\cite{ryffel2018} & \redcross & \ForestGreencheck & \ForestGreencheck & \redcross & \redcross & \redcross\\
SecureML~\cite{mohassel2017} & \redcross & \ForestGreencheck & \redcross & \ForestGreencheck & \redcross & \redcross\\
XONN~\cite{sadegh2019} & \ForestGreencheck & N/A & \redcross & \redcross & \redcross & \redcross\setcounter{rownum}{1}\\
\midrule
\multicolumn{7}{l}{\textit{Three parties}}\\
\midrule
ABY3~\cite{mohassel2018} & \redcross & N/A & \redcross & \ForestGreencheck & \redcross & \redcross\\
Astra~\cite{chaudhari2019} & \redcross & \ForestGreencheck & \redcross & \ForestGreencheck & \redcross & \redcross\\
Blaze~\cite{patra2020} & \redcross & \ForestGreencheck & \redcross & \ForestGreencheck & \redcross & \redcross\\
CrypTFlow~\cite{kumar2020} & \redcross & N/A & \redcross & \redcross & \ForestGreencheck & \redcross\\
CryptGPU$^\ddagger$~\cite{cryptgpu} & \redcross & \redcross & \ForestGreencheck & \ForestGreencheck & \ForestGreencheck & \ForestGreencheck\\ 
Falcon~\cite{wagh2021} & \ForestGreencheck & N/A & \redcross & \ForestGreencheck & \ForestGreencheck & \redcross\\
SecureNN~\cite{wagh2018securenn} & \redcross & N/A & \redcross & \ForestGreencheck & \redcross & \redcross\setcounter{rownum}{1}\\
\midrule
\multicolumn{7}{l}{\textit{Four parties}}\\
\midrule
FLASH~\cite{byali2020} & \ForestGreencheck & N/A & \redcross & \ForestGreencheck & \redcross & \redcross\\
Trident~\cite{rachuri2019} & \ForestGreencheck & N/A & \redcross & \ForestGreencheck & \redcross & \redcross\setcounter{rownum}{1}\\
\midrule
\multicolumn{7}{l}{\textit{Arbitrary number of parties}}\\
\midrule
\crypten{} (ours) & \redcross & \redcross$^\mathsection$ & \ForestGreencheck & \ForestGreencheck & \ForestGreencheck & \ForestGreencheck\\
\bottomrule
\end{tabularx} 
}
\caption{Overview of secure MPC frameworks for machine learning and their properties. $^\dagger$We define a framework to be general-purpose if it supports all of the following layers: linear, convolution, rectified linear unit (ReLU), max-pooling, and batch normalization. $^\ddagger$We note that CryptGPU was developed \emph{on top of} \crypten{}, hence, it inherits many features from \crypten{}. $^\mathsection$Future versions of \crypten{} will support Beaver triple generation without requiring a trusted third party.}
\label{tab:framework_comparison}
\end{table}

\end{document}